\documentclass[11pt]{article}

\PassOptionsToPackage{dvipsnames,table}{xcolor}
\usepackage[final]{acl}

\usepackage{times}
\usepackage{latexsym}

\usepackage[T1]{fontenc}

\usepackage[utf8]{inputenc}

\usepackage{microtype}

\usepackage{inconsolata}

\usepackage{graphicx}

\usepackage{tabularx}
\usepackage{booktabs}
\usepackage{array}
\usepackage{makecell}
\usepackage{subcaption}
\definecolor{seaRed}{HTML}{B04A2A} 
\definecolor{seaBlue}{HTML}{2B5AA8} 
\definecolor{seaGreen}{HTML}{2C8D57}
\usepackage{url}

\usepackage{siunitx}
\definecolor{langEn}{HTML}{F4CCCC}   
\definecolor{langTl}{HTML}{FFD966}  
\definecolor{langId}{HTML}{FFE599}   
\definecolor{langKm}{HTML}{B6D7A8}   
\definecolor{langLo}{HTML}{A2C4C9}   
\definecolor{langMs}{HTML}{CFE2F3}   
\definecolor{langTa}{HTML}{E06666}   
\definecolor{langVi}{HTML}{93C47D}   
\definecolor{langZh}{HTML}{6D9EEB}   
\definecolor{langTh}{HTML}{CEC6E8}   
\definecolor{langMy}{HTML}{EAD1DC}   

\newcommand{\lang}[2]{\colorbox{#1}{\textsf{#2}}}
\usepackage{enumitem}
\usepackage{multirow}
\usepackage{float}

\usepackage{amssymb}
\usepackage[normalem]{ulem}

\newcommand{\metaused}{\ensuremath{\checkmark}}
\newcommand{\metafiltered}{\ensuremath{\checkmark^{\mathrm{F}}}}
\newcommand{\metaunused}{\textcolor{gray!65}{\ensuremath{\checkmark^{\dagger}}}}
\newcommand{\metana}{--}

\definecolor{revisionblue}{RGB}{0,0,0}
\usepackage{amsmath,amsfonts,bm}

\title{\textsc{SEA-SpeechBench}: A Large-Scale Multitask Benchmark for Speech Understanding Across Southeast Asia}

\author{
 \textbf{Jingyi Liao\textsuperscript{1,3}}\thanks{Equal contribution.},
 \textbf{Wenyu Zhang\textsuperscript{2}}\footnotemark[1]\thanks{Work done at Institute for Infocomm Research (now Institute of Advanced Intelligence and Computing), A*STAR.},
 \textbf{Zhuohan Liu\textsuperscript{1}},
 \textbf{Yingxu He\textsuperscript{1}},
 \textbf{Geyu Lin\textsuperscript{1}},
 \textbf{Xunlong Zou\textsuperscript{1}},
 \\
 \textbf{Shuo Sun\textsuperscript{1}},
 \textbf{Syed Ali Redha Alsagoff\textsuperscript{3}}\footnotemark[2],
 \textbf{Ai Ti Aw\textsuperscript{1}}
 \\\\
 \textsuperscript{1}Institute of Advanced Intelligence and Computing, A*STAR \\
 \textsuperscript{2}Center for AI Safety, \\
 \textsuperscript{3}Nanyang Technological University \\
\small{
  \textbf{Correspondence:}
  \href{mailto:liao_jingyi@a-star.edu.sg}{liao\_jingyi@a-star.edu.sg},
  \href{mailto:wen.projectz@gmail.com}{wen.projectz@gmail.com}
}
}

\begin{document}
\maketitle
\begin{abstract}
The rapid advancement of audio and multimodal large language models has unlocked transformative speech understanding capabilities, yet evaluation frameworks remain predominantly English-centric, leaving Southeast Asian (SEA) languages critically underrepresented. We introduce \textsc{SEA-SpeechBench}, to the best of our knowledge, the first large-scale multitask benchmark that evaluates speech understanding in 11 SEA languages through 97,194 samples across 99 evaluation sets and 597 hours of curated audio data. Our benchmark comprises 9 diverse tasks across 3 categories: speech processing (automatic speech recognition, speech translation, spoken question answering), paralinguistic analysis (emotion, gender, age, speaker recognition), and temporal understanding, a novel dimension featuring timestamped content queries and temporal localization within extended audio sequences up to 3 minutes. We implement multilingual prompting in both native SEA languages and English to reflect user interactions with audio-language models. 
Evaluation of leading open-source and proprietary systems reveals marked performance gaps. Across all models, performance remains underwhelming on temporal understanding, emotion recognition, and speech translation. 
Prompting in low-resource languages such as Burmese and Tamil lags behind English by up to 41 percentage points.
Our findings expose critical model limitations and underscore the need for inclusive model development. 
The \textsc{SEA-SpeechBench} benchmark is available at \href{https://zwenyu.github.io/SEA-SpeechBench/}{\texttt{zwenyu.github.io/SEA-SpeechBench}}.
\end{abstract}

\section{Introduction}

\begin{figure*}[!t]
  \centering
  \includegraphics[width=0.98\linewidth]{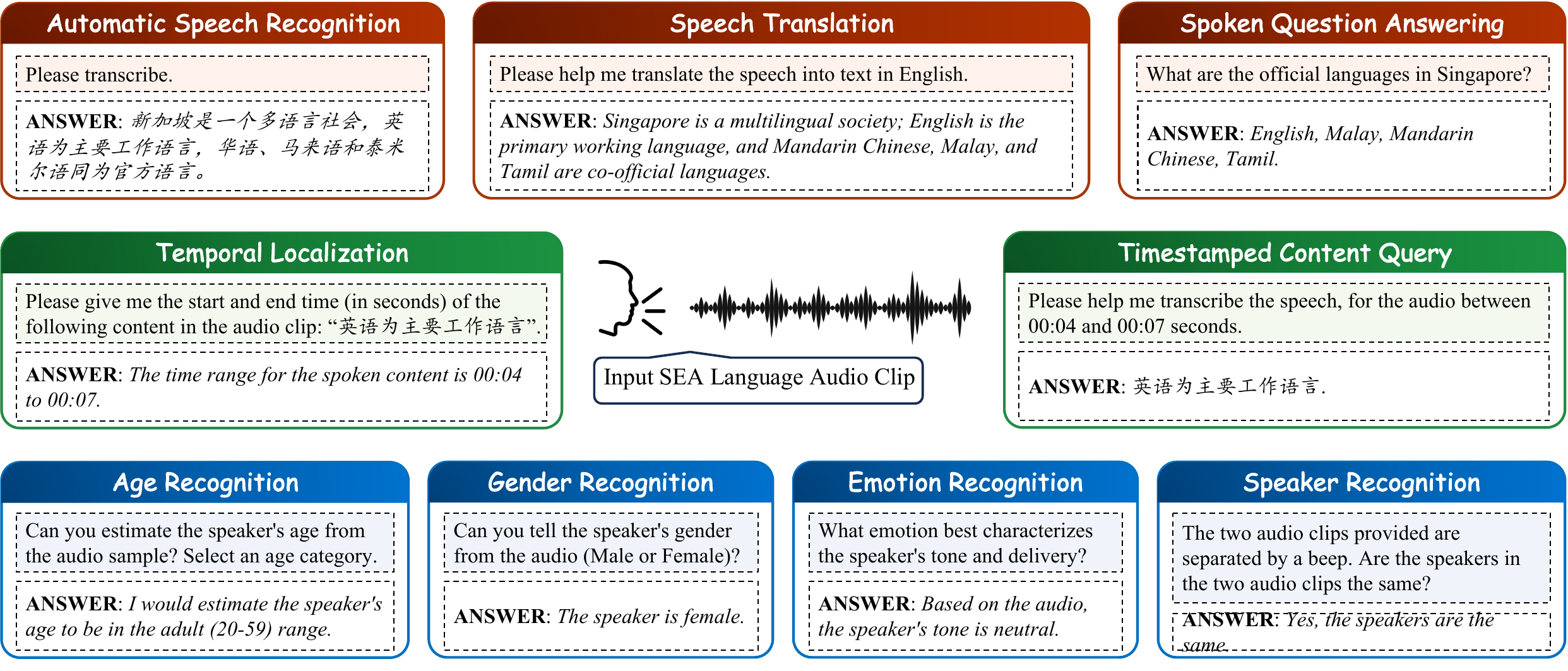}
  \caption{Overview of \textsc{SEA-SpeechBench} task suite. The suite covers nine tasks in three categories.
\textcolor{seaRed}{\rule{6pt}{6pt}}\;\textbf{Speech processing},
\textcolor{seaBlue}{\rule{6pt}{6pt}}\;\textbf{Paralinguistic}, and
\textcolor{seaGreen}{\rule{6pt}{6pt}}\;\textbf{Temporal reasoning}.}
  \label{fig:tasksuite}
  \vspace{-0.5cm}
\end{figure*}

Recent advancement in audio large language models (AudioLLMs) has led to transformative applications in voice assistants, transcription, accessibility technologies, and multimodal reasoning~\citep{NExT-GPT, gemini25,SpeechGPT}. Despite these advances, research in speech understanding has been disproportionately concentrated on high-resource languages, particularly English and a small number of European and East Asian languages~\citep{SUPERB_benchmark,voxpopuli,aishell}. While recent benchmarking efforts~\citep{mmau, audiobench, airbench} have made significant strides in evaluating audio-language models across diverse tasks and modalities, they universally overlook Southeast Asian (SEA) languages, leaving an entire linguistic region underexplored despite representing over \textit{650 million speakers} worldwide.

Developing comprehensive benchmarks for SEA languages also presents unique technical challenges. The region's speech landscape is characterized by extraordinary linguistic diversity, rich tonal and phonetic structures, and substantial resource disparities across languages: factors that create evaluation complexities absent from English-centric benchmarks. Many SEA languages operate in low-resource contexts with limited annotated data and sparse digital representation, making robust evaluation both methodologically challenging and critically important for equitable technological development.
While recent initiatives, such as MERaLiON \citep{merlion} which targets Singapore's multilingual context, Typhoon2-Audio \citep{typhoon2} which focuses on Thai, and SeaLLMs-Audio \citep{SeaLLMs-Audio} which extends capabilities to selected SEA languages, have begun to build general-purpose speech-language models for the SEA region, these efforts remain limited in both scope and linguistic coverage. Crucially, they lack comprehensive evaluation frameworks necessary to systematically assess capabilities across the full spectrum of Southeast Asian speech understanding tasks. Meanwhile, fragmented data collection efforts across research groups have yielded heterogeneous SEA datasets~\citep{seacrowd, vietnam_celeb, IndoWaveSentiment, malcsc}, but without unified frameworks for task definitions, prompting, and normalization, standardized comparison remains difficult.

In this work, we introduce to our knowledge, the first ever comprehensive benchmark for speech understanding in Southeast Asian languages, designed to evaluate the capabilities of general-purpose speech-text LLMs. We focus on the official languages of Southeast Asian countries, as in Table~\ref{tab:country-official-langs}, to balance broad regional coverage and practical relevance. 

\begin{table}[tb]
\centering
\resizebox{0.95\linewidth}{!}{%
\begin{tabular}{l|l|l}
\toprule
\textbf{Country} & \textbf{Official language(s)} & \textbf{ISO 639-1 Code(s)} \\
\midrule
Singapore   & English, Malay, Mandarin Chinese, Tamil & en, ms, zh, ta \\
Malaysia    & Malay (Bahasa Melayu)            & ms \\
Indonesia   & Indonesian (Bahasa Indonesia)    & id \\
Philippines & Filipino / Tagalog, English      & tl, en \\
Thailand    & Thai                             & th \\
Cambodia    & Khmer (Cambodian)                & km \\
Lao PDR     & Lao                              & lo \\
Myanmar     & Burmese (Myanmar)                & my \\
Vietnam     & Vietnamese                       & vi \\
\bottomrule
\end{tabular}}
\caption{Southeast Asian countries and their official language(s).%
\;\;Note: Several countries recognize additional regional or minority languages at sub-national levels; this table lists state-level official languages corresponding to the language codes used in our benchmark.}
\label{tab:country-official-langs}
\vspace{-0.5cm}
\end{table}

Our benchmark encompasses 11 SEA languages with over 597 hours of audio data, curated from existing sources and synthesized into new tasks and datasets through systematic processing.
The benchmark spans 9 diverse tasks across 3 broad categories: speech processing, paralinguistics, and temporal reasoning. 
We introduce two tasks that evaluate models’ capacity for temporal reasoning and localization within audio streams, addressing a previously unexplored dimension in audio LLM evaluation. These tasks test models’ ability to navigate time-dependent information and extract content from specific temporal locations. The complete task suite is illustrated in Figure \ref{fig:tasksuite}. To better reflect authentic usage scenarios, we evaluate each task using both English and native language text prompts.

\textbf{The main contributions of this paper are threefold. }First, we present \textsc{SEA-SpeechBench}, to our knowledge, the first ever large-scale multitask benchmark that systematically evaluates speech processing, paralinguistic analysis, and temporal reasoning across Southeast Asian languages. It comprises a total of 99 evaluation sets with more than 97,000 audio samples. Second, we introduce temporal reasoning tasks that assess models’ ability to reason about time-dependent information in extended audio sequences. Finally, we comprehensively evaluate both open-source and proprietary models, offering critical insights into their strengths, limitations, and areas for further research.

\section{\textsc{SEA-SpeechBench} Evaluation Suite}

\subsection{Task Suite}
\label{sec:tasks}


\textsc{SEA-SpeechBench} comprises 9 core tasks across 3 categories: 
\textit{speech processing}, \textit{paralinguistic analysis}, and \textit{temporal reasoning}. All tasks require models to produce textual responses given an audio input and a text query.
\begin{table*}[!htbp]
\centering
\resizebox{0.75\linewidth}{!}{
\begin{tabular}{l|c|c|c|c|c|c}
\toprule
\textbf{Task} & \textbf{Languages} & \textbf{\#Datasets} & \textbf{\#Samples} & \textbf{Total L (h)} & \textbf{Min L (s)} & \textbf{Max L (s)}\\
\midrule
\textbf{ASR} &
\makecell[c]{%
  \lang{langEn}{en} \lang{langTl}{tl} \lang{langId}{id}
  \lang{langKm}{km} \lang{langLo}{lo} \\
  \lang{langMs}{ms} \lang{langMy}{my} \lang{langTa}{ta} \lang{langVi}{vi}
  \lang{langZh}{zh} \lang{langTh}{th}%
}
& 33 & 26,863 & 52.88 & 0.47 & 30.00 \\
\midrule
\textbf{ST} &
\makecell[c]{%
  \lang{langEn}{en} \lang{langTl}{tl} \lang{langId}{id}
  \lang{langKm}{km} \lang{langLo}{lo} \\
  \lang{langMs}{ms} \lang{langMy}{my} \lang{langVi}{vi} \lang{langTh}{th}
}
& 9 & 7,189 & 26.46 & 3.06 & 30.00 \\
\midrule
\textbf{SQA} &
\lang{langEn}{en}\ \lang{langZh}{zh}\ \lang{langId}{id}\ \lang{langTh}{th} \ \lang{langVi}{vi}
& 7 & 5,462 & 40.57 & 20.00 & 30.00 \\
\midrule
\textbf{ER} &
\lang{langZh}{zh}\ \lang{langId}{id}\ \lang{langTh}{th}\ \lang{langEn}{en}\ \lang{langTa}{ta}
& 7 & 5,356 & 5.33 & 0.12 & 29.86 \\
\midrule
\textbf{GR} &
\makecell[c]{%
  \lang{langZh}{zh} \
    \lang{langId}{id} \
    \lang{langTh}{th} \
    \lang{langEn}{en} \
    \lang{langTa}{ta} \ \\
    \lang{langVi}{vi} \
    \lang{langKm}{km} \
    \lang{langMy}{my}
}
& 16 & 13,599 & 22.02 & 0.12 & 29.90 \\
\midrule
\textbf{AgeR} &
\lang{langZh}{zh} \
\lang{langTh}{th} \
\lang{langEn}{en} \
\lang{langTa}{ta} \
\lang{langVi}{vi}
& 5 & 4,608 & 6.55 & 0.58 & 20.78 \\
\midrule
\textbf{SpkR} &
\lang{langZh}{zh} \
\lang{langTh}{th} \
\lang{langEn}{en} \
\lang{langTa}{ta} \
\lang{langVi}{vi} \
\lang{langMy}{my}
& 8 & 7,827 & 19.72 & 2.10 & 30.38 \\
\midrule
\textbf{TCQ} &
\lang{langZh}{zh} \
\lang{langEn}{en} \
\lang{langTh}{th} \
\lang{langId}{id} \
\lang{langVi}{vi}
& 7 & 13,145 & 211.98 & 20.00 & 180.00 \\
\midrule
\textbf{TLoc} &
\lang{langZh}{zh} \
\lang{langEn}{en} \
\lang{langTh}{th} \
\lang{langId}{id} \
\lang{langVi}{vi}
& 7 & 13,145 & 211.98 & 20.00 & 180.00 \\
\midrule
\textbf{Total} & -- & \textbf{99} & \textbf{97,194} & \textbf{597.49} & -- & --\\
\bottomrule
\end{tabular}
}
\caption{Summary of curated datasets for \textsc{SEA-SpeechBench}. For multilingual datasets, each language-specific sub-dataset is counted separately in \#Datasets.}
\label{tab:dataset-summary}
\vspace{-0.3cm}
\end{table*}

\emph{Speech processing} covers three fundamental capabilities: Automatic Speech Recognition (ASR), Speech Translation (ST) from Southeast Asian languages to English, and Spoken Question Answering (SQA) based on SEA speech inputs.

\emph{Paralinguistic analysis} examines vocal cues beyond linguistic content. It includes four tasks: Emotion Recognition (ER), \textcolor{revisionblue}{which classifies each utterance into a closed nine-class emotion set (neutral, happy, angry, sad, surprised, disgusted, fearful, frustrated, disappointed)}; Gender Recognition (GR), which predicts gender from voice characteristics; Age Recognition (AgeR), which categorizes speakers as teens (10–19), adults (20–59), or seniors (60–100); and Speaker Recognition (SpkR), which determines whether two clips belong to the same speaker.

\emph{Temporal reasoning} 
\textcolor{revisionblue}{ treats audio as a searchable temporal space, mirroring how users interact with long-form recordings through queries such as \textit{``what was said at this time''} and \textit{``when was this mentioned''}. We introduce two complementary tasks that probe both directions of this temporal--content mapping. \textbf{Timestamped Content Query (TCQ)} tests the time-to-content direction: given an interval $[t_s, t_e]$, the model extracts the speech content within that window, evaluating localized retrieval. \textbf{Temporal Localization (TLoc)} tests the content-to-time direction: given a textual query, the model predicts the exact time span $\hat{y}=[\hat{t}_s,\hat{t}_e]$ where it appears, evaluating boundary detection and temporal grounding. 
}

The first two categories use short clips ($\leq 30$s) to align with most current model input limits. As real-world audio applications increasingly involve longer recordings where users require temporal navigation, \textsc{SEA-SpeechBench} introduces the temporal reasoning tasks to assess model ability to perform reasoning and localization over extended sequences of up to 3 minutes.

\subsection{Data Curation and Processing}

\paragraph{Source selection and annotation use.}

\textsc{SEA-SpeechBench} is curated from public and community speech corpora, \textcolor{revisionblue}{where source inclusion is task-specific, as shown in Table ~\ref{tab:source-selection}. For ASR and ST, we retain sources with reliable utterance-level transcription or translation references under short-clip conditions. Long-form YouTube-sourced corpora~\cite{yodas} are accordingly admitted only for SQA, TCQ, and TLoc construction, where filtered transcript fragments suffice, but excluded from ASR evaluation that requires gold utterance-level references.} For AgeR, GR, ER, and SpkR, we use only source-provided metadata, such as age, gender, emotion, and speaker identity, and do not infer missing attributes post hoc. Speaker labels are additionally used to enforce speaker-disjoint splits and construct same/different speaker pairs. For SQA, TCQ, and TLoc, we require transcript metadata and, for temporal tasks, timestamp or alignment information. Full source-language-task mapping, dataset details, metadata availability, and licenses are reported in Appendix~\ref{app:dataset}.

\begin{table*}[t]
\centering
\scriptsize
\setlength{\tabcolsep}{3pt}
\begin{tabular}{p{0.12\textwidth} p{0.54\textwidth} p{0.28\textwidth}}
\toprule
Task(s) & Source datasets & Required annotation for inclusion into task \\
\midrule
ASR / ST & FLEURS \citep{fleurs}, Commonvoice \citep{commonvoice}, OpenSLR \citep{OpenSLR1, OpenSLR2, OpenSLR3}, Bloom-Speech \cite{bloomspeech}, Thai Elderly Speech \citep{thaielderly}, THAI SER \citep{thaiser}, ASR-SMalDuSC \citep{SMalDuSC}, VietMed \citep{VietMed}, Bud500 \citep{Bud500}, LOTUS \citep{lotus}, SG Streets \citep{sg_streets}, ASR-SgpCCSC \cite{sgpccsc}, ASR-MALCSC \citep{malcsc}, MIG \cite{mig}, ESD \cite{esd} & Utterance-level transcripts or segment-level translations; short clips within 30 seconds. \\
AgeR / GR / ER & FLEURS \citep{fleurs}, Commonvoice \citep{commonvoice}, OpenSLR \citep{OpenSLR1, OpenSLR2, OpenSLR3}, M3ED \citep{m3ed}, EmoTa \citep{emota}, Thai Elderly Speech \citep{thaielderly}, THAI SER \citep{thaiser}, ASR-SMalDuSC \citep{SMalDuSC}, SG Streets \citep{sg_streets}, ESD \citep{esd}, TEC \cite{tec}, SFDUSC \citep{sfdusc_ph}, Vietnam-Celeb \cite{vietnam_celeb}, IndoWaveSentiment \cite{IndoWaveSentiment} & Age, gender, or emotion labels; short clips within 30 seconds.  \\
SpkR & ESD \citep{esd}, EmoTa \citep{emota}, SFDUSC \citep{sfdusc_ph}, MIG \cite{mig}, Thai Elderly Speech \citep{thaielderly}, THAI SER \cite{thaiser}, ASR-SMalDuSC \citep{SMalDuSC}, VoxVietnam-O \citep{voxvietnam} & Speaker labels; short clips within 30 seconds.  \\
SQA & SG Streets \citep{sg_streets}, YODAS2 \citep{yodas}, ASR-SgpCCSC \cite{sgpccsc} & Paragraph-level transcripts; speech-transcript pairs pass CTC alignment, language, and content filters; generated QA pairs pass human audit; short clips within 30 seconds.  \\
TCQ / TLoc & \citep{sg_streets}, YODAS2 \citep{yodas}, ASR-SgpCCSC \cite{sgpccsc} & Paragraph-level transcripts; speech-transcript pairs pass CTC alignment, language, and content filters; segment-level timestamps; long clips up to 3 minutes. \\
\bottomrule
\end{tabular}
\caption{\textcolor{revisionblue}{Task-specific source selection criteria. Full statistics, metadata availability, and licenses are in Appendix~A.}}
\vspace{-0.3cm}
\label{tab:source-selection}
\end{table*}

\begin{figure*}[!htbp]
  \centering

  \begin{subfigure}[t]{0.35\linewidth}
    \centering
    \includegraphics[width=\linewidth]{\detokenize{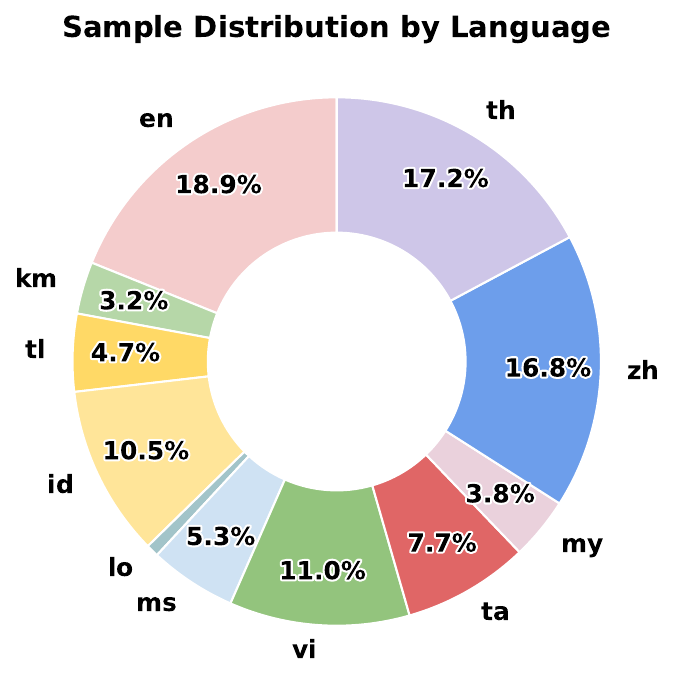}}
    \caption{Sample distribution by language. Colors match Table~\ref{tab:dataset-summary} language tags.}
    \label{fig:lang-hours-pie}
  \end{subfigure}\hfill
  \begin{subfigure}[t]{0.34\linewidth}
    \centering
    \includegraphics[width=\linewidth]{\detokenize{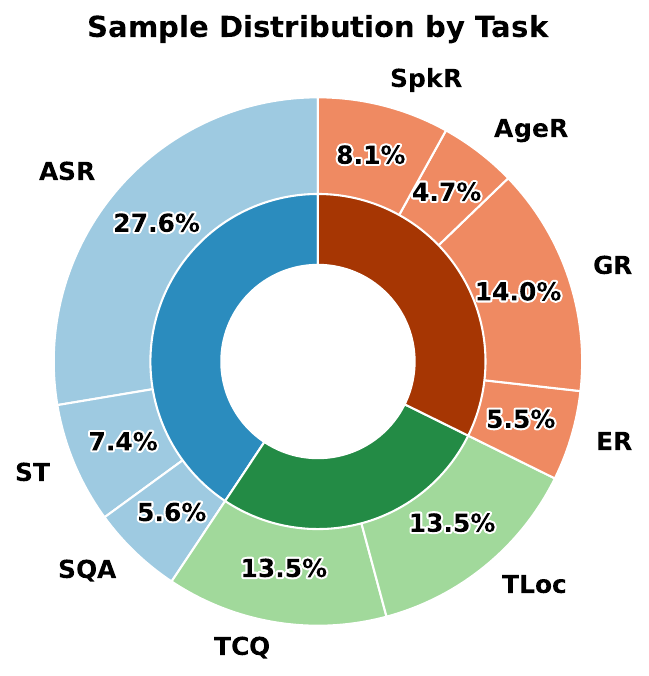}}
    \caption{Sample distribution by task.}
    \label{fig:task-pie}
  \end{subfigure}
  \begin{subfigure}[t]{0.29\linewidth}
    \centering
    \includegraphics[width=\linewidth]{\detokenize{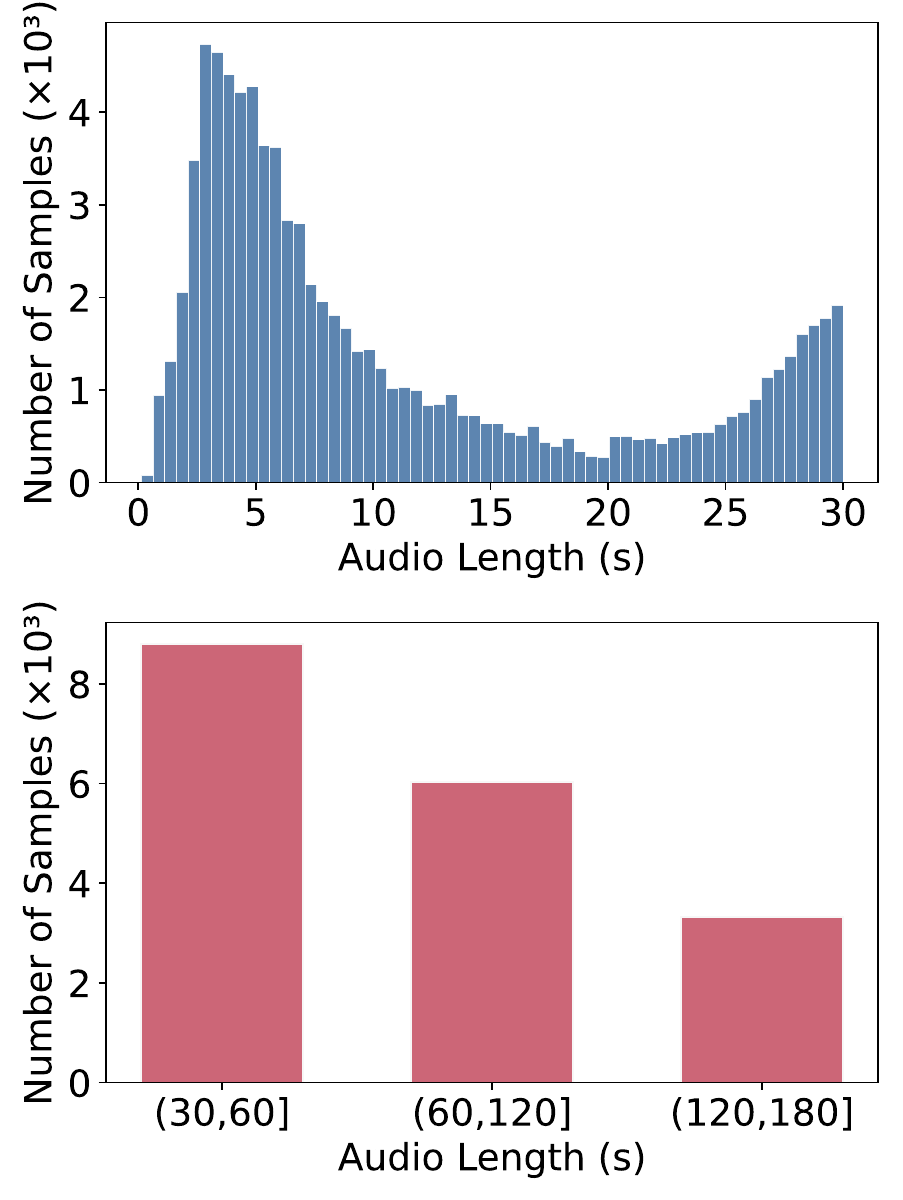}}
    \caption{Audio duration distribution overview.}
    \label{fig:length}
  \end{subfigure}
  \caption{\textsc{SEA-SpeechBench} composition: (a) language distribution by audio hours, (b) task distribution by sample count, and (c) audio duration distribution. (Upper: short audios ($\leq30$s) with fine-grained histogram. Lower: long audios ($>30$s) aggregated into three duration ranges.) 
  }
  \label{fig:overview-data}
  \vspace{-0.5cm}
\end{figure*}

\noindent\textbf{Dataset Statistics:} 
As shown in Table~\ref{tab:dataset-summary}, \textsc{SEA-SpeechBench} is a comprehensive benchmark comprising of over 97,000 samples across 9 tasks and 11 Southeast Asian languages.


\noindent\textbf{Language and Task Distribution:} Figure~\ref{fig:overview-data} reveals the linguistic and task composition of our benchmark. The language distribution (Figure~\ref{fig:lang-hours-pie}) demonstrates substantial coverage across Southeast Asian languages, with English (18.9\%), Thai (17.2\%) and Chinese (16.8\%) representing the largest segments, followed by Vietnamese (11.0\%) and Indonesian (10.5\%). 
We also include low-resource languages such as Khmer and Lao to ensure representation of the region's full linguistic spectrum. 
Figure~\ref{fig:task-pie} shows balanced coverage across evaluation task families.

\noindent\textbf{Audio Length Distribution:} We cover both short and long audios to enable more comprehensive and realistic evaluation. As shown in Figure \ref{fig:length}, our benchmark encompasses a broad distribution of utterances spanning 0–30 seconds, with natural concentration around sentence-length segments for traditional tasks. For temporal understanding evaluation, we systematically curate extended recordings through strategic segmentation of long-form datasets, creating stratified duration bins of 30–60s, 60–120s, and 120–180s, totaling over 10,000 samples for long audios.

\noindent\textbf{Data Processing:} 
We applied uniform data processing across all datasets and synthesized evaluation sets for tasks lacking suitable existing data. 

We unify all data sources into a consistent triplet format {audio context, text instruction, text answer} with paired audio/text fields. All audios are converted to a standard representation by downmixing to mono when needed, resampling to 16 kHz, applying peak normalization to control amplitude variations, and storing the waveform in an OGG/OPUS container to ensure consistent decoding and comparable inference conditions across datasets. For evaluation, we sample a fixed-size test set per language and duration bin (up to 1,000 instances) with a fixed random seed for reproducibility. Also, we enforce a speaker-disjoint split: segments associated with the same speaker are assigned to a single split only, so no speaker appears in both train and test. Within each split, candidates are filtered by our quality criteria and then sampled to match the target size, with a minimum-count safeguard for low-resource bins.

For long-form recordings requiring timestamped queries (e.g., TCQ/TLoc), segments are built from contiguous speech within the same recording and may span multiple utterances as long as the inter-utterance gap is at most 5 seconds. Segments are terminated when the duration exceeds a target bin (e.g., 60–120s), the gap between utterances exceeds 5 seconds, or utterance-level alignment confidence is low (CTC log-probability $<-2$).
To ensure sufficient semantic content, we apply a speech-density filter (words-per-second for space-delimited languages; characters-per-second otherwise) and remove templated or boilerplate phrases. We further filter using CTC-confidence thresholds at both utterance and segment levels, and perform language identification using the Lingua Python language-detection library on the concatenated transcript, retaining segments only when the predicted ISO-639-1 code matches the target language. Finally, we deduplicate segments by transcript to remove near-duplicates.
More data processing and synthesis details are provided in Appendix~\ref{app:processing}.

\subsection{Metrics}~\label{sec:metric}

\textcolor{revisionblue}{Table~\ref{tab:task-metrics} presents each task's output format and primary evaluation metric. ASR and TCQ use WER/CER against gold transcripts; ST uses BLEU and chrF; TLoc uses a deterministic span-overlap F1 over predicted and reference timestamps. For GR, SpkR, and AgeR, we canonicalize each model output to a candidate class label through deterministic string matching against bilingual label lexicons and compute macro-F1, which is robust to the demographic skew present in AgeR (Appendix~\ref{app:label-balance}). Macro-F1 weighs all classes equally and is therefore more appropriate than accuracy or micro-F1 under such imbalance. For SQA and ER, it is challenging to reliably canonicalize free-form outputs, and we therefore employ an LLM-based judge adapted from \cite{audiobench}. The SQA judge assigns a 0--5 quality score, linearly rescaled to $s_{\text{SQA}} = 20 \times \text{score}$. The ER judge canonicalizes outputs to candidate emotion labels and renders binary correctness, aggregated as \textbf{accuracy}. Per-dataset ER label distributions are near-uniform (Appendix~\ref{app:label-balance}). Detailed judging prompts are in Appendix~\ref{app:judge_prompt}.}


\begin{table}[tb]
\centering
\resizebox{0.95\linewidth}{!}{%
\begin{tabular}{l|l|l}
\toprule
\textbf{Task} & \textbf{Expected Output} & \textbf{Metric(s)} \\
\midrule
\textbf{ASR} & Text transcript & \textbf{WER/CER} \\
\textbf{ST} & Text translation & \textbf{BLEU, chrF}  
\\
\textbf{SQA} & Short textual answer & \textbf{Scaled judge score {$s_{SQA}$}} $^{\dagger}$
\\
\midrule
\bf AgeR & Age bin $\in$ \{teens, adults, seniors\} &  \bf Macro-F1 \\
\bf ER   & Emotion label (Appendix \ref{app:er-label-map}) & \bf Judge-based Acc$^\dagger$ \\
\bf GR   & Gender label $\in$ \{male, female\} &  \bf Macro-F1 \\
\bf SpkR & Speaker identity match or mismatch &  \bf Macro-F1 \\
\midrule
\textbf{TCQ}& Text transcript at given time & \textbf{WER/CER} \\
\textbf{TLoc}& Time span $[t_s,t_e]$ & \textbf{$F1$ score} \\
\bottomrule
\end{tabular}}
\caption{\textcolor{revisionblue}{Outputs and metrics by task. $^{\dagger}$ \emph{Judge-provided metrics}: scores are produced by a LLM serving as an external judge.}}
\vspace{-0.3cm}
\label{tab:task-metrics}
\end{table}


\noindent\textbf{Assessment of Model Judge:}
To assess alignment with human judgment, we conduct an expert human evaluation for the two judge-dependent tasks in \textsc{SEA-Speechbench} (SQA and ER), covering three candidate model judges: Gemma-3-27B-Instruct~\cite{gemma3}, Qwen3-Omni-30B-A3B-Instruct~\cite{qwen3}, and GPT-4.1~\cite{gpt4}. We analyze judge--human agreement separately for English prompts and SEA-language prompts. For each task and language, we randomly selected 100 model responses and collected both model-judge scores and human-evaluator scores from native-speaker annotators. Alignment is quantified using Matthews correlation coefficient (MCC) for ER (binary correctness) and Spearman rank correlation ($\rho$) for SQA (ordinal 0--5 scale). Table~\ref{tab:agreement} shows the quantitative results.


\begin{table}[tb]
  \centering
  \resizebox{0.95\linewidth}{!}{%
    \begin{tabular}
    {l|c|c|c|c}
        \toprule
          & \multicolumn{1}{|c}{\makecell{MCC \\ (Eng)}} & \multicolumn{1}{|c}{\makecell{$\rho$ \\ (Eng)}} & \multicolumn{1}{|c}{\makecell{MCC \\ (SEA)}} & \multicolumn{1}{|c}{\makecell{$\rho$ \\ (SEA)}} \\
\midrule
Qwen3-Omni-30B-A3B-Instruct & 0.81 & 0.80 &  0.76  &  0.37 \\
Gemma-3-27B-Instruct & \bf 0.98 & \underline{0.82} &  \bf 0.97  &  \underline{0.66} \\
GPT-4.1 & \bf 0.98  & \bf 0.84  & \bf 0.97 & \bf 0.76 \\
\bottomrule
    \end{tabular}%
    }
    \caption{Consistency between LLM-judged scores and human evaluator scores for English and SEA prompts. Alignment is quantified using Matthews correlation coefficient (MCC) for binary classification tasks and Spearman rank correlation ($\rho$) for ordinal ratings.}
  \label{tab:agreement}%
  \vspace{-0.3cm}
\end{table}%

For ER under English prompts, Gemma-3-27B-Instruct and GPT-4.1 both exhibit strong and comparable alignment with human annotations. For SQA under SEA-language prompts, GPT-4.1 achieves substantially higher judge--human correlation than Gemma-3 ($\rho = 0.76$ vs.\ $\rho = 0.66$, averaged across SEA languages). Balancing evaluation performance and computational cost, we adopt a hybrid judging strategy: GPT-4.1 serves as the judge for SQA, while Gemma-3-27B-Instruct is deployed for ER.
 




\section{Evaluations}

\subsection{Evaluation Setting}
We conduct extensive evaluation across several state-of-the-art open-source audio/multimodal LLMs with multiple size variants ranging from 2B to 30B parameters:
MERaLiON-2~\citep{merlion}, SeaLLMs-audio~\citep{SeaLLMs-Audio}, Phi-4-multimodal-instruct~\citep{phi4}, Qwen2-Audio-Instruct~\citep{Qwen2-Audio}, Qwen2.5-Omni~\citep{Qwen2.5-Omni}, Gemma-3n-it~\citep{gemma3n}, Voxtral~\citep{Voxtral}, Kimi-audio~\citep{kimi}, and Qwen3-Omni and its thinking model~\citep{qwen3}. We additionally evaluate a cascaded model setting: Whisper-large-v3-turbo~\citep{whisper} with Qwen3.6 35B A3B~\citep{qwen3.6} on speech translation task. All models are evaluated using their official released checkpoints with recommended inference configurations to ensure fair comparison. To establish comprehensive performance baselines, we also evaluate two leading commercial models: Gemini 2.5 Flash~\citep{gemini25} and GPT-4o~\citep{gpt4o}. For GPT-4o, we employ the specialized Whisper-based transcription API for ASR tasks and the general audio understanding model for all other tasks, ensuring optimal performance.


\subsection{Results and Insights}
\textcolor{revisionblue}{
Throughout Tables~\ref{tab:asr_lang_avg_eng}--\ref{tab:tcq-tloc-by-duration}, \textbf{bold} marks the best model within each group (open-source / commercial), and \uline{underline} marks the global best across all evaluated models.
}

\subsubsection{ASR Capability}
We report the ASR performance of each model across different languages under English prompts in Table~\ref{tab:asr_lang_avg_eng}. Gemini 2.5 Flash has the best ASR performance overall, and MERaLiON-2 has the best performance amongst open-source models. The instruction-tuned Qwen3-Omni model is also highly competitive, achieving the best results on several languages, including Vietnamese, Indonesian, and Thai. We note that state-of-the-art performance on Tamil and Burmese is particularly low at more than 0.25 WER and CER, respectively.

\begin{table*}[t]
\centering
\resizebox{0.78\linewidth}{!}{
\begin{tabular}{l|c|ccccccccccc|c}
\toprule
\textbf{Model} & \textbf{Size} & \textbf{en} & \textbf{tl} & \textbf{vi} & \textbf{id} & \textbf{ta} & \textbf{th} & \textbf{zh} & \textbf{km} & \textbf{lo} & \textbf{ms} & \textbf{my} & \textbf{Avg.} \\
\midrule
Gemma-3n-it   & 2B       & 1.32  & 0.42  & 3.77  & 0.17  & 1.20  & 3.03  & 1.55  & 3.92  & 1.42  & 1.06  & 2.81 & 1.88 \\
Qwen2.5-omni & 3B  & \bf \uline{0.07}  & 0.60  & 0.27  & 0.15  & 1.36  & 0.12  & \bf \uline{0.05}  & 2.97  & 1.39  & 0.23  & 2.01 & 0.84 \\
MERaLiON-2  & 3B    & \bf \uline{0.07}  & 0.20  & 0.35  & 0.11  & 0.43  & 0.08  & 0.11  & 1.72  & 0.86  & 0.17  & 1.19 & 0.48 \\
Voxtral  & 3B   & 0.21  & 2.40  & 0.93  & 0.36  & 1.20  & 0.57  & 0.45  & 1.72  & 1.29  & 0.76  & 5.05 & 1.36 \\
Gemma-3n-it   & 4B       & 0.76  & 0.24  & 2.71  & 0.51  & 0.45  & 0.32  & 0.69  & 1.71  & \bf \uline{0.15}  & 1.98  & 2.98 & 1.14 \\
SeaLLMs-Audio  & 7B      & 0.38  & 1.03  & 0.44  & 0.28  & 1.52  & 0.05  & 0.34  & 1.07  & 1.01  & 0.61  & 1.20 & 0.72 \\
Phi-4 & 5.6B  & 0.09  & 5.10  & 2.92  & 2.76  & 1.93  & 2.98  & 0.10  & 3.11  & 2.45  & 3.15  & 2.13 & 2.43 \\
Qwen2-Audio-it & 7B  & 0.14  & 1.88  & 1.03  & 0.74  & 1.48  & 1.21  & 0.21  & 1.08  & 1.08  & 1.02  & 1.17 & 1.00 \\
Qwen2.5-omni & 7B  & \bf \uline{0.07}  & 0.55  & 0.25  & 0.10  & 1.34  & 0.54  & \bf \uline{0.05}  & 3.08  & 2.63  & 0.50  & 5.49 & 1.33 \\
Kimi-Audio    & 7B  & 0.25  & 3.42  & 15.86 & 0.58  & 4.46  & 2.60  & \bf \uline{0.05}  & 5.27  & 4.52  & 5.92  & 7.99 & 4.63 \\
MERaLiON-2  & 10B   & \bf \uline{0.07}  & \bf 0.18  & 0.23  & 0.09  & \bf 0.38  & 0.10  & 0.09  & \bf 0.77  & 0.39  & \bf 0.13  & \bf 0.83 & \bf 0.30 \\
Qwen3-omni-Thinking  & 30B & 0.11 & 0.51 & 0.20 & 0.06 & 0.86 & 0.04 & 0.13 & 5.71 & 1.06 & 0.29 & 7.21 & 1.47 \\
Qwen3-omni-it  & 30B & 0.09 & 0.42 & \bf 0.19 & \bf 0.05 & 0.44 & \bf \uline{0.03} & 0.06 & 1.78 & 0.49 & 0.29 & 2.27 & 0.56 \\
\midrule
Gemini 2.5 Flash & -    & \textbf{0.11}  & \bf \uline{0.13}  & \bf \uline{0.12}  & \bf \uline{0.03}  & \bf \uline{0.27}  & \bf 0.05  & 0.18  & \bf \uline{0.13}  & \bf 0.16  & \bf \uline{0.11}  & \bf \uline{0.31} & \bf \uline{0.15} \\
GPT-4o & -  & 0.13  & 0.14  & 0.21  & 0.05  & 0.41  & 0.06  & \bf 0.09  & 0.30  & 0.38  & 0.17  & 0.66 & 0.24 \\
\bottomrule
\end{tabular}
}
\caption{\textbf{English-prompt ASR results averaged by language.} Metrics reported are raw WER/CER (e.g., $0.07$ corresponds to $7\%$): CER for languages without explicit word boundaries, WER otherwise (\S\ref{sec:metric}). Lower is better.}
\vspace{-0.3cm}
\label{tab:asr_lang_avg_eng}
\end{table*}

\subsubsection{Speech Processing and Paralinguistics}
Tables~\ref{tab:sp} and~\ref{tab:para} present the results of additional speech processing and paralinguistic tasks. Results are reported under both English and SEA prompts. 
ST performance is limited: open-source models top out at 21 BLEU and 45--47 chrF, trailing commercial models at 48--54 chrF. 
The two metrics agree on overall ranking, with one notable divergence: Gemini 2.5 Flash reaches 53.94 chrF under SEA prompts against only 18.89 BLEU, indicating that its translations preserve character-level content even when surface word choices diverge from references. For paralinguistics tasks, emotion recognition is especially challenging. Apart from Kimi-Audio, no model achieved scores above 25. Even the best-performing Kimi-Audio did not score above 50.

\begin{table}[h]
\centering
\resizebox{\columnwidth}{!}{%
\begin{tabular}{l|c|cc|cc|cc}
\toprule
\multirow{2}{*}{\bf Model} & \multirow{2}{*}{\bf Size}
  & \multicolumn{2}{c}{\bf ST (BLEU)}
  & \multicolumn{2}{c}{\bf ST (chrF)}
  & \multicolumn{2}{c}{\bf SQA} \\
\cmidrule(lr){3-4} \cmidrule(lr){5-6} \cmidrule(lr){7-8}
 & & ENG & SEA & ENG & SEA & ENG & SEA \\
\midrule
Gemma-3n-it           & 2B   & 8.97  & 8.72  & 30.98 & 30.76 & 73.35 & 56.06 \\
Qwen2.5-omni          & 3B   & 7.60  & 6.27  & 33.00 & 28.46 & 78.42 & 73.14 \\
MERaLiON-2            & 3B   & 7.56  & 7.42  & 28.36 & 29.22 & 66.68 & 60.19 \\
Voxtral               & 3B   & 19.98 & 18.15 & 43.52 & 36.86 & 82.32 & 79.06 \\
Gemma-3n-it           & 4B   & 10.98 & 13.58 & 33.14 & 39.87 & 79.24 & 78.89 \\
Phi-4                 & 5.6B & 3.04  & 0.32  & 17.10 & 4.87  & 64.69 & 47.36 \\
SeaLLMs-Audio         & 7B   & 10.74 & 10.11 & 32.44 & 28.40 & 76.42 & 78.38 \\
Qwen2-Audio-it        & 7B   & 4.54  & 3.20  & 23.37 & 19.46 & 65.82 & 57.95 \\
Qwen2.5-omni          & 7B   & 7.91  & 8.06  & 35.87 & 35.35 & 71.60 & 75.57 \\
Kimi-Audio            & 7B   & 3.36  & 7.71  & 18.02 & 24.35 & 69.62 & 63.49 \\
MERaLiON-2            & 10B  & 17.75 & 19.52 & 42.97 & 45.56 & 82.00 & 82.05 \\
Qwen3-omni-Thinking   & 30B  & 12.97 & 11.95 & 37.70 & 37.59 & 84.93 & 85.28 \\
Qwen3-omni-it         & 30B  & 15.28 & 13.55 & 41.56 & 39.66 & \textbf{85.77} & \textbf{85.62} \\
Whisper \& Qwen3.6 & - & \bf 20.83 & \bf 20.96	 & \bf 44.87 & \bf 47.29 & - & - \\
\midrule
Gemini 2.5 Flash      & --   & 16.86 & 18.89 & 45.13 & \textbf{\uline{53.94}} & \textbf{\uline{92.28}} & \textbf{\uline{86.21}} \\
GPT-4o                & --   & \textbf{\uline{21.24}} & \textbf{\uline{21.39}} & \textbf{\uline{48.30}} & 48.36 & 86.66 & 82.44 \\
\bottomrule
\end{tabular}%
}
\caption{Results on speech processing tasks (ST, SQA) under English (ENG) and native SEA (SEA) prompts. ST is evaluated with BLEU (word-level $n$-gram overlap) and chrF (character-level $F$-score); SQA is reported as a scaled judge score. Higher is better for all metrics.}
\vspace{-0.3cm}
\label{tab:sp}
\end{table}


\begin{table}[htbp]
\centering
\vspace{-0.1cm}
\resizebox{\linewidth}{!}{%
\begin{tabular}{l|c|cc|cc|cc|cc}
\toprule
\multirow{2}{*}{\textbf{Model}} 
& \multirow{2}{*}{\textbf{Size}} 
& \multicolumn{2}{c|}{\textbf{AgeR}} 
& \multicolumn{2}{c|}{\textbf{ER}} 
& \multicolumn{2}{c|}{\textbf{GR}} 
& \multicolumn{2}{c}{\textbf{SpkR}} \\
\cmidrule(lr){3-4}\cmidrule(lr){5-6}\cmidrule(lr){7-8}\cmidrule(lr){9-10}
 & & \small ENG & \small SEA 
   & \small ENG & \small SEA 
   & \small ENG & \small SEA 
   & \small ENG & \small SEA \\
\midrule
Gemma-3n-it   & 2B   & 26.82 & 22.70 & 12.21 & 13.17 & 11.49 & 10.05 & 37.16 & 32.07 \\
Qwen2.5-omni      & 3B   & 29.78 & 24.75 & 13.45 &  9.95 & 49.85 & 37.30 & 25.22 & 31.47 \\
MERaLiON-2        & 3B   & 33.14 & 28.16 & 23.99 & 18.73 & 41.44 & 36.48 & 43.45 & 29.91 \\
Voxtral      & 3B   & 37.26 & 28.49 & 10.62 &  5.35 & 39.78 & 20.41 & 42.84 & 38.67 \\
Gemma-3n-it   & 4B   & 32.41 & 29.97 & 12.46 & 13.93 & 35.41 & 23.54 & 33.01 & 34.13 \\
Phi-4 & 5.6B & 26.28 & 28.42 & 20.87 &  9.20 & 49.36 & 31.06 & 36.43 & 32.73  \\
SeaLLMs-Audio  & 7B   & 29.77 & 18.08 & 12.34 &  9.17 & 37.07 & 32.77 & 43.50 & 31.05 \\
Qwen2-Audio-it & 7B & 17.23 & 15.39 & 24.47 & 19.36 & 91.89 & 66.24 & 41.93 & 31.66 \\
Qwen2.5-omni      & 7B   & 15.23 & 18.28 & 16.33 & 10.45 & 65.03 & 49.17 & 15.55 & 18.61 \\
Kimi-Audio        & 7B   & 35.42 & 34.09 & \textbf{\uline{36.86}} & \textbf{\uline{42.55}} & 92.73 & \textbf{71.36} & 46.58 & 29.61 \\
MERaLiON-2        & 10B  & 32.99 & 32.46 & 18.73 & 20.34 & 57.50 & 45.93 & 48.36 & 37.07\\
Qwen3-omni-Thinking  &  30B & 38.10 & \textbf{\uline{36.78}} & 15.06 & 10.34 & 94.77 & 69.34 & 52.35 & 43.29 \\
Qwen3-omni-it  &  30B & \textbf{\uline{39.56}} & 36.70 & 15.94 & 11.31 & \textbf{\uline{95.78}} & 65.43 & \textbf{56.15} & \textbf{44.28} \\
\midrule
Gemini~2.5~Flash  & --   & \textbf{36.93} & \textbf{36.33} & \bf 19.50 & 16.79 & \textbf{92.54} & \textbf{\uline{82.54}} & \textbf{\uline{58.01}} & \textbf{\uline{51.54}} \\
GPT\mbox{-}4o & -- & 36.91 & 34.27 & 17.00 & \bf 19.50 & 59.40 & 46.78 & 14.49 & 19.04 \\
\bottomrule
\end{tabular}}
\caption{Results on paralinguistic tasks (AgeR, ER, GR, SpkR) under English and SEA prompts, respectively.}
\label{tab:para}
\vspace{-0.3cm}
\end{table}

\begin{table*}[htb]
\centering
\begin{subtable}{0.76\textwidth}
\centering
\vspace{-0.2cm}
\resizebox{\linewidth}{!}{%
\begin{tabular}{l| c | c c c c | c c c c}
\toprule
 &  & \multicolumn{4}{c|}{\textbf{TCQ (WER/CER $\downarrow$)}} & \multicolumn{4}{c}{\textbf{TLoc (F1 Score $\uparrow$)}} \\
\cmidrule(lr){3-6}\cmidrule(l){7-10}
\textbf{Model} & \textbf{Size} & \textbf{0–30 s} & \textbf{30–60 s} & \textbf{60–120 s} & \textbf{120–180 s} & \textbf{0–30 s} & \textbf{30–60 s} & \textbf{60–120 s} & \textbf{120–180 s} \\
\midrule
SeaLLMs-Audio & 7B & 5.48 & - & - & - & 11.57 & - & - & - \\
Qwen2-Audio-it & 7B & 5.56 & - & - & - & 33.30 & - & - & - \\
Qwen2.5-omni & 3B & 8.74 &	9.20 & - & - & 30.49 &	18.21 & - & - \\
Gemma-3n-it & 2B & 7.32	& 7.49 & - & - & 11.82	& 8.27 & - & - \\
Gemma-3n-it & 4B & 6.76 & 7.38 & - & - & 13.25 & 8.72 & - & - \\
Phi-4 & 5.6B & 20.67 &	14.90 &	16.25 & - & 12.97	& 6.25	& 3.64 & - \\
MERaLiON-2 & 3B & 4.77	& 8.21 & 	11.27 & - & 18.82 & 10.28 & 5.14 & - \\
Qwen2.5-omni & 7B & 5.49  & 6.58  & 9.85 & - & \bf \uline{35.74} & \bf \uline{19.98} & \bf \uline{11.32} & - \\
Kimi-Audio & 7B & 14.00 &	19.07 &	24.04 &	42.78 & 14.49	 & 8.61 & 3.70 & 3.00 \\
Voxtral & 3B & 4.74 & 7.74 & 	12.86 & 	22.13 & 17.85	& 9.87 & 3.71 & 2.45 \\
MERaLiON-2 & 10B & 5.12& 	9.48 & 14.72 & 17.67 & 22.22 & 12.37 & 6.40 & 4.53 \\
Qwen3-omni-Thinking  &  30B & 1.28 & 1.35 & 2.17 & 3.12 & 10.80 & 5.76 & 2.32 & 1.96 \\
Qwen3-omni-it  &  30B & \bf \uline{1.18} & \bf \uline{1.22} & \bf \uline{1.34} & \bf \uline{1.86} & 11.40 & 5.90 & 2.43 & 1.28 \\
\midrule
Gemini 2.5 Flash & - & \bf 2.41	& \bf 2.64	& \bf 9.44	& \bf 4.09 & 11.66 & 7.70	& 5.57 & 5.30 \\
GPT-4o & - & 5.06	& 	6.82	& 	7.38	& 	8.59 & \bf 26.47 & \bf 17.83 & \bf 8.53 & \bf \uline{5.38} \\
\bottomrule
\end{tabular}}
\caption{\textbf{English-prompt TCQ and TLoc results (\%) by duration.} A dash (–) indicates audio lengths for which the model is unable to perform inference.}
\label{tab:tcq-tloc-by-duration}
\end{subtable}
\hfill
\begin{subtable}{0.19\textwidth}
\centering
\vspace{-0.2cm}
\includegraphics[width=\linewidth]{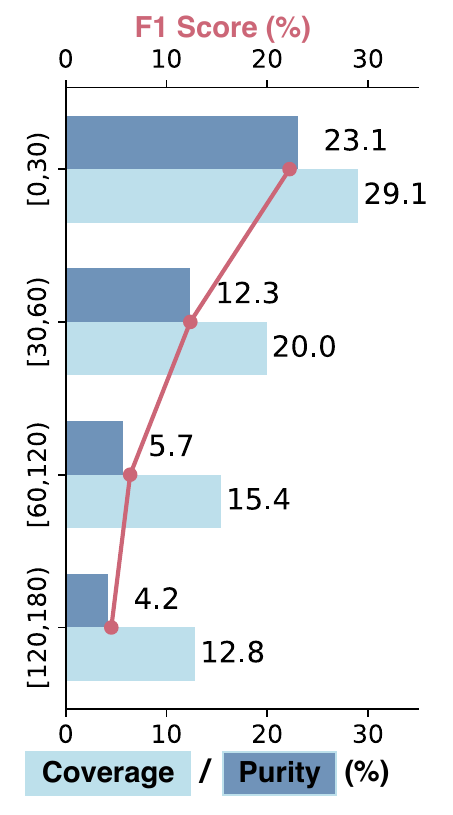}
\caption{TLoc metrics for MERaLiON-2-10B.}
\label{fig:coverage}
\end{subtable}
\caption{Temporal reasoning performance by duration.}
\label{tab:coverage}
\vspace{-0.5cm}
\end{table*}

\subsubsection{Temporal Reasoning}\label{sec:time}
In this section, we provide detailed analysis of temporal reasoning capabilities, stratifying performance across four duration bins: [0,30), [30,60), [60,120), [120,180) seconds as shown in Table~\ref{tab:tcq-tloc-by-duration}. Figure~\ref{fig:coverage} presents coverage ($C$), purity ($P$), and $F_1$ of TLoc task across audio-duration ranges, using MERaLiON-2-10B as a case study.

\noindent{\color{revisionblue}\textbf{Difficulty Design.}
Temporal reasoning difficulty scales with the temporal search space,
which we control through four duration bins (0--30s, 30--60s, 60--120s,
120--180s). Longer audios enlarge the candidate space for boundary localization
and place stronger demands on long-range memory.}

\noindent\textbf{Metrics Analysis.} First, we observe systematic over-coverage in temporal reasoning across models. Notably, the instruction-tuned Qwen3-Omni model achieves exceptionally strong TCQ performance across all duration bins, substantially outperforming other models in WER/CER. In TCQ, models frequently produce content that extends beyond the queried time window. In TLoc, as demonstrated in Figure~\ref{fig:coverage}, coverage consistently exceeds purity across all durations, which is a pattern we observe in most evaluated models. This asymmetry indicates weak temporal boundary localization and alignment, reflecting a recall-seeking strategy that favors longer spans, and thus higher coverage, at the cost of precision. These motivate finer-grained temporal grounding, boundary-aware training objectives, and decoding constraints that penalize span over-coverage. 

\noindent\textbf{Constraints on Audio Length.} As shown in Table~\ref{tab:tcq-tloc-by-duration}, only a select subset of models sustains inference availability across all duration bins. 
Both TCQ and TLoc exhibit performance degradation with increasing duration: errors accumulate over longer contexts, manifesting as boundary drift and truncation, exposing current architectural limits.



\subsubsection{Effect of Prompt Language}\label{sec:prompt_analysis}

We systematically investigate how prompt language choice: English versus native SEA language, affects model performance across our benchmark.

\begin{figure}[!t]
  \centering
  \includegraphics[width=0.95\linewidth]{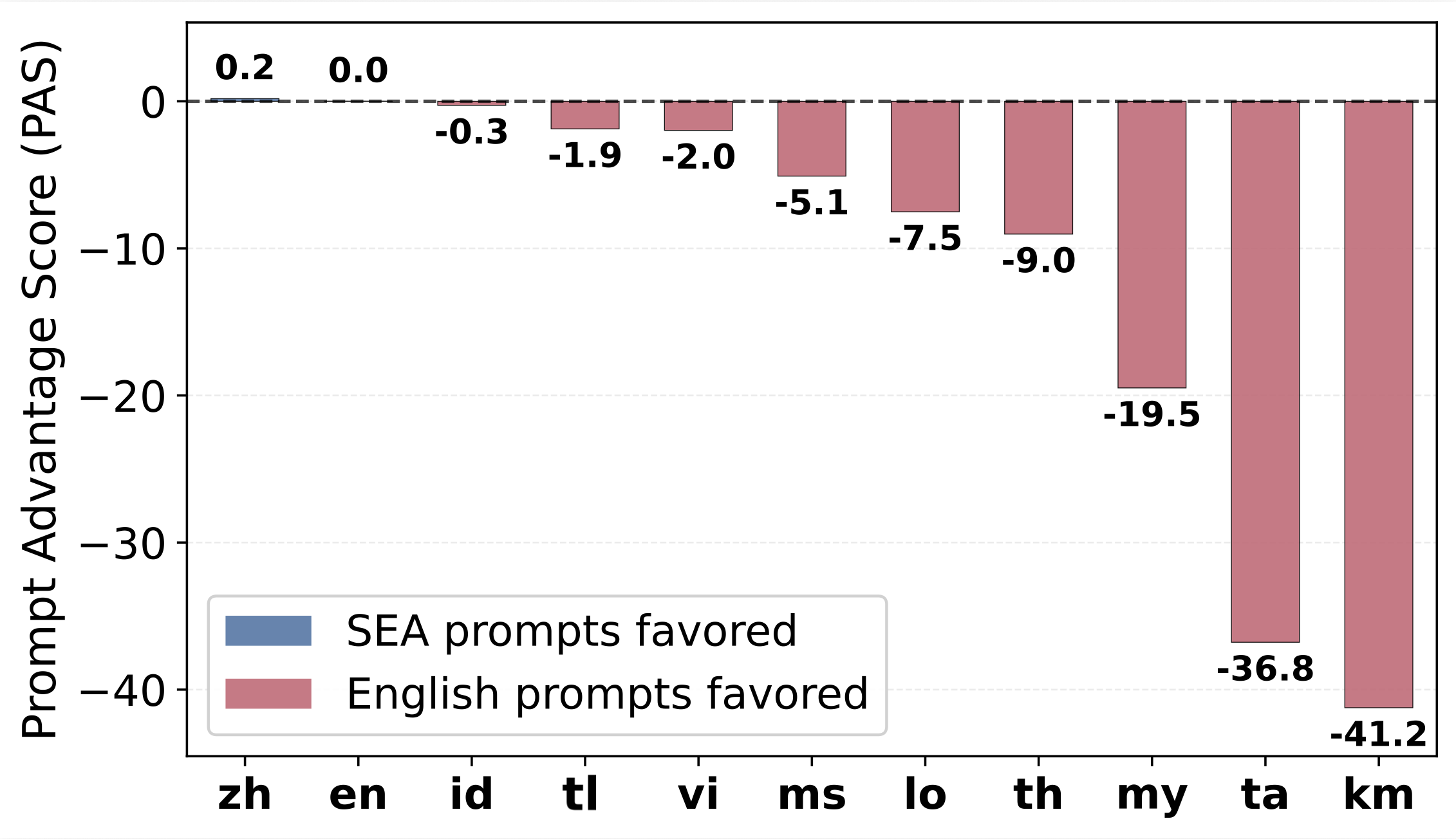}
  \vspace{-0.15cm}
  \captionof{figure}{Cross-linguistic prompt sensitivity measured by Prompt Advantage Score (PAS).}
  \vspace{-0.5cm}
  \label{fig:prompt_rank}
\end{figure}

Figure~\ref{fig:prompt_rank} shows a cross-linguistic hierarchy in prompt sensitivity. We define a Prompt Advantage Score (PAS) to quantify this effect, with details in Appendix~\ref{app:pas}. Higher PAS values indicate stronger local language prompt advantage, while negative scores suggest English prompt superiority for that particular language.
Chinese, English, and Indonesian cluster near zero (+0.2, 0.0, $-$0.3), with prompt-language choice essentially neutral. The remaining SEA languages display increasing English preference across two distinct clusters: moderate disadvantages for Filipino, Vietnamese, Malay, Lao, and Thai (ranging from $-$1.9 to $-$9.0), and severe English advantages for Myanmar, Tamil, and Khmer ($-$19.5 to $-$41.2).

\begin{figure}[tb]
  \centering
  \includegraphics[width=0.85\linewidth]{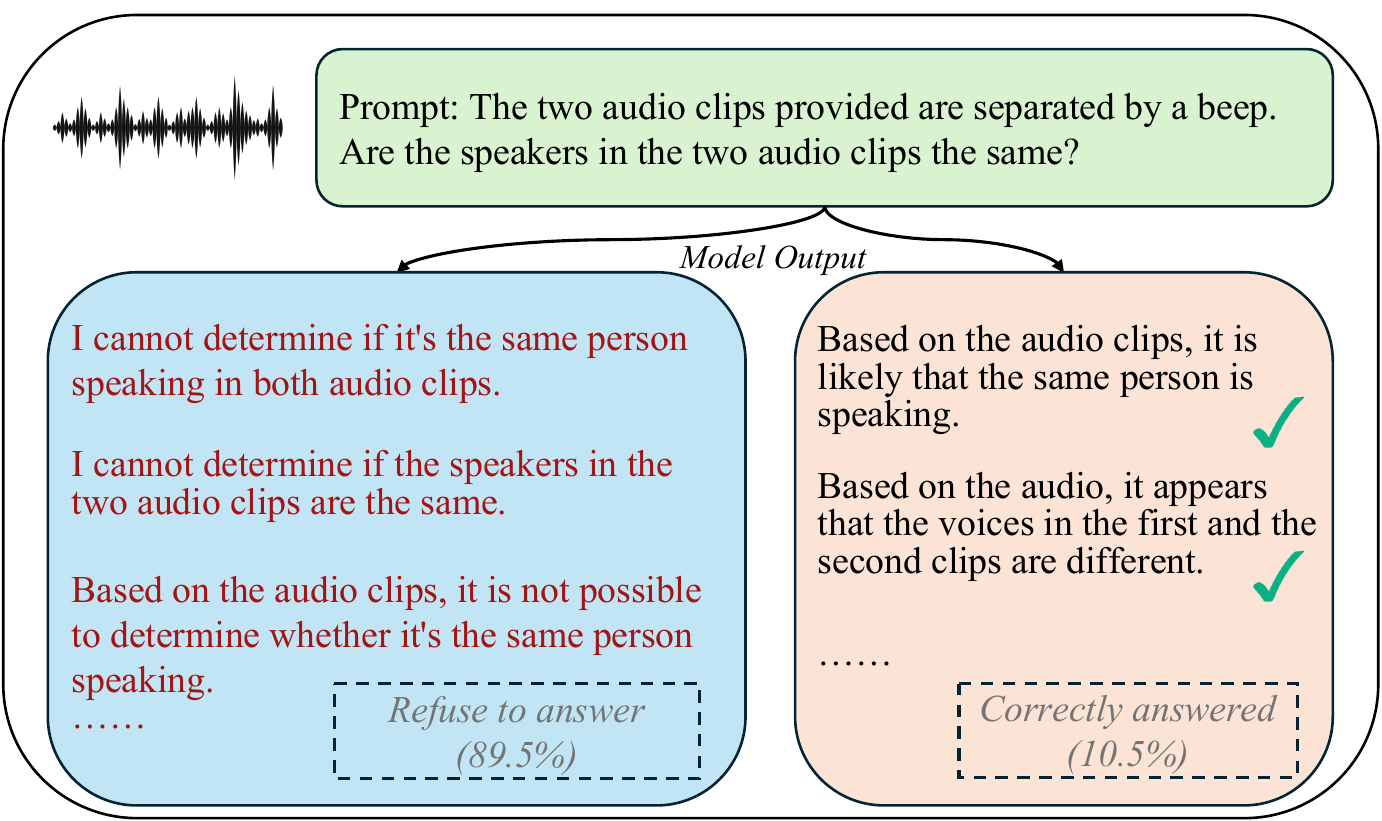}
  \captionof{figure}{Speaker recognition failure examples in GPT-4o responses.}
  \label{fig:failure}
  \vspace{-0.6cm}
\end{figure}

\subsubsection{Refusal Behavior}\label{sec:refuse}

From Table~\ref{tab:para}, GPT-4o attains only 14.49\% accuracy on speaker recognition. Inspecting the error breakdown shows that these “errors” are refusals rather than wrong predictions. As illustrated in Figure~\ref{fig:failure}, GPT-4o refuses to answer 89.5\% of the queries. When the model does answer for 10.5\% of the queries, it is consistently correct (non-refusal accuracy = 100\%). 
This pattern emerges on our self-constructed SpkR dataset, 
which is likely out-of-distribution for GPT-4o. We hypothesize that the high refusal rate may stem from built-in safety policies 
against speaker identification, which the model treats as a privacy-sensitive task. This is consistent with the fact that when GPT-4o does answer, its non-refusal accuracy reaches 100\%, suggesting deliberate abstention rather than task incomprehension. 
By contrast, Qwen2.5-Omni-7B also attains low SpkR performance with frequent refusals, but its accepted responses contain nontrivial mistakes, pointing to weaker calibration and label 
grounding rather than abstention alone.





\section{Related Works}

\paragraph{Audio/Multimodal LLMs and Benchmarks.}
Recent audio-language models connect speech encoders to LLMs via lightweight adaptors~\citep{SALMONN,SpeechGPT,SpeechLLaMA}, share speech-text token spaces~\citep{AudioPaLM,SpiritLM}, or introduce stronger multimodal and temporal modeling in systems such as Phi-4, Qwen2.5-Omni, Voxtral, GPT-4o, and Gemini~\citep{phi4,Qwen2.5-Omni,Voxtral,gpt4o,gemini25}. Evaluation has similarly expanded from speech recognition to instruction following and reasoning through benchmarks such as SUPERB, AudioBench, AirBench, MMAU, MMAU-Pro, IFEval-Audio, and domain-specific audio evaluations~\citep{SUPERB_benchmark,audiobench,airbench,mmau,mmaupro,IFEval,2025cmi}. However, existing models and benchmarks remain predominantly English-centric, with limited coverage of Southeast Asian languages. Regional efforts such as MERaLiON and SeaLLMs-Audio~\citep{merlion,SeaLLMs-Audio} cover only part of the linguistic space. \textsc{SEA-SpeechBench} addresses this gap by evaluating 11 SEA languages across 9 tasks under unified prompting and normalization.

\section{Conclusion}

We present \textsc{SEA-SpeechBench}, the first comprehensive benchmark for evaluating speech understanding across 11 Southeast Asian languages, comprising 97,000+ samples across 9 tasks in speech processing, paralinguistics, and temporal reasoning. Our standardized framework enables reproducible, cross-linguistic comparisons through unified normalization and bilingual prompting.

Evaluation of leading commercial and open-source systems exposes systematic weaknesses: performance collapses on long audio (temporal brittleness), English prompts consistently outperform native languages (linguistic inequity), and tasks such as temporal reasoning, emotion recognition, and speech translation remain far below usability thresholds. These findings underscore persistent scalability and generalization gaps. By surfacing these limitations, \textsc{SEA-SpeechBench} seeks to establish a rigorous baseline for developing temporally robust, linguistically inclusive, and practically deployable speech technologies for Southeast Asia’s diverse communities.

\section{Limitations and Risks}
Data availability remains a key limitation of this work. Despite extensive efforts to collect and curate data from diverse public and community-driven sources, coverage of certain dialects and task categories remains limited due to the fundamental scarcity of high-quality, annotated resources. This constraint restricts the breadth of dialectal and task-level evaluation that can be reliably supported. Future work may explore data-efficient learning paradigms, weakly supervised or synthetic data generation, and collaborative data collection initiatives to further extend coverage across underrepresented dialects and tasks.

Risk of test set contamination is another limitation. This risk is inherently difficult to control because most evaluated models are with undisclosed training data and curation pipelines.

\section*{Acknowledgments}
This research is supported by the National Research Foundation, Singapore under its National Large Language Models Funding Initiative. Any opinions, findings, conclusions, or recommendations expressed in this material are those of the author(s) and do not reflect the views of the National Research Foundation, Singapore.

We are grateful to Aye Phyu Phyu Aung, Nattadaporn Lertcheva, Rodel Miguel, and Manh Cuong Nguyen for their helpful discussions and assistance in understanding Southeast Asian language features for data processing, as well as Hardik Sailor and Qiongqiong Wang for their support testing the emotion recognition tasks.

\bibliography{custom}

\appendix

\section{Dataset Catalog}\label{app:dataset}

Table~\ref{tab:catalog} lists the source datasets used in our SEA-SpeechBench, along with their tasks, languages, audio lengths and licensing information.

{\color{revisionblue}
\paragraph{Metadata availability and usage.}
Table~\ref{tab:metadata-availability} summarizes which metadata fields are provided by each source dataset and which are used in ~\textsc{SEA-Speechbench}. A field is marked as used if it supports evaluation labels or references, filtering, pairing, timestamp construction, or split enforcement; source-provided but unused fields are marked separately. This distinction reflects our task-specific curation policy: metadata is used only when it is reliable and directly supports the target evaluation.

For YODAS2, transcripts are used after CTC, language-identification, and content filtering for SQA, TCQ, and TLoc construction, but are not used as gold ASR references. This is because YODAS2 is collected from YouTube long-form audio and its source transcripts are not uniformly reliable at the utterance level: they may contain loose segmentation, alignment noise, incomplete local utterances, or transcript fragments that are sufficient for filtered temporal query construction but not suitable as gold sentence-level ASR references. We therefore use YODAS2 selectively: a permissive filtered subset provides enough continuous speech for long-form SQA/TCQ/TLoc construction, while high-confidence aligned utterances are used only for localized temporal queries. We exclude it from standard ASR evaluation to avoid mixing gold transcriptions with weakly aligned or automatically derived transcripts.

Other unused fields are excluded for task-specific reasons. Age metadata from multiple datasets is not used for AgeR because it does not provide a sufficiently balanced three-way age distribution. Transcripts from emotion datasets such as THAI-SER, IndoWaveSentiment, M3ED, and EmoTa are not used as ASR references because these datasets often contain repeated or templated text spoken with different prosodic or emotional styles. For sources with richer auxiliary annotations, such as English and Chinese ESD dataset, we retain only the fields needed for ASR, ER, and SpkR construction and leave redundant metadata such as gender unused.

\paragraph{Release plan.}
~\textsc{SEA-Speechbench} will be released as an \textbf{evaluation benchmark}: the fixed sampled evaluation set constitutes the official test split and serves as the stable artifact for reproducible comparison. Each evaluation sample carries a unique identifier that maps back to its original source dataset, allowing users to locate evaluation samples within the source data and derive their own training splits without leakage. When source licenses permit redistribution, we additionally provide the corresponding processed audio; when redistribution is restricted or unclear, we provide manifest-only access along with evaluation scripts, prompts, and reconstruction instructions, so users obtain the original data from the source provider and reconstruct the evaluation subset locally. This protocol enables reproducible evaluation without redistributing data in conflict with source licenses.
}

\begin{table*}[!ht]
  \centering
  \resizebox{0.95\linewidth}{!}{%
    \begin{tabular}{lllcccl}
      \toprule
      & \textbf{Languages}
      & \textbf{Tasks}
      & \multicolumn{1}{c}{\textbf{Total L (hr)}}
      & \multicolumn{1}{c}{\textbf{Min L (s)}}
      & \multicolumn{1}{c}{\textbf{Max L (s)}}
      & \multicolumn{1}{l}{\textbf{License}} \\
      \midrule
    \textbf{Commonvoice} ~\citep{commonvoice}  & zh, vi, th, id, ta, en & GR, AGE, ASR & 22.66 & 0.58  & 20.78 & MPL 2.0 \\
    \textbf{FLEURS}~\citep{fleurs} & \makecell[l]{zh, vi, th, my, ms, \\lo, km, id, tl} & ASR, ST, GR & 57.79 & 3.06  & 30.00    & CC-BY 4.0 \\
    \makecell[l]{\textbf{OpenSLR}~\citep{OpenSLR1}\\~\citep{OpenSLR2,OpenSLR3}} & ta, my, km & GR, ASR & 6.33  & 1.71  & 17.15 & CC-BY-SA 4.0 \\
    \textbf{Bloom-Speech}~\citep{bloomspeech} & tl, my, en & ASR   & 0.84 & 0.47  & 29.46 & \multicolumn{1}{l}{CC BY-NC 4.0} \\
    \textbf{Thai Elderly Speech}~\citep{thaielderly} & th    & ASR, GR, SpkR & 5.66  & 2.04  & 27.06 & \multicolumn{1}{l}{CC-BY-SA 4.0} \\
    \textbf{LOTUS}~\citep{lotus} & th    & ASR   & 1.22  & 1.84  & 29.78 & \multicolumn{1}{l}{CC-BY-NC-SA 4.0} \\
    \textbf{THAI SER}~\citep{thaiser} & th    & SpkR, GR, ER & 6.22  & 0.60 & 29.86 & \multicolumn{1}{l}{CC-BY-SA 4.0} \\
    \textbf{SG Streets$^\dagger$}~\citep{sg_streets} & en    & \makecell[l]{TLoc, TCQ, SQA, \\ASR, GR} & 6.55$^\dagger$  & 0.80   & 59.97 & \multicolumn{1}{l}{NA} \\
    \textbf{VietMed}~\citep{VietMed} & vi    & ASR   & 1.74  & 2.00     & 12.00    & \multicolumn{1}{l}{MIT} \\
    \textbf{VoxVietnam-O}~\citep{voxvietnam} & vi    & SpkR  & 2.43  & 1.42  & 29.68 & \multicolumn{1}{l}{CC} \\
    \textbf{Bud500}~\citep{Bud500} & vi    & ASR   & 0.71  & 1.01  & 5.48  & \multicolumn{1}{l}{Apache-2.0} \\
    \textbf{Vietnam-Celeb}~\citep{vietnam_celeb} & vi    & GR    & 2.14  & 0.84  & 29.66 & \multicolumn{1}{l}{CC-BY-4.0} \\
    \textbf{ASR-SMalDuSC}~\citep{SMalDuSC} & ms    & ASR, SpkR, GR & 8.51  & 2.64  & 29.22 & \multicolumn{1}{l}{CC BY-NC-ND 4.0} \\
    \textbf{ASR-MALCSC}~\citep{malcsc} & ms    & ASR   & 0.74  & 0.61  & 18.16 & \multicolumn{1}{l}{CC BY-NC-ND 4.0} \\
    \textbf{IndoWaveSentiment}~\citep{IndoWaveSentiment} & id    & GR, ER & 0.52  & 3.00     & 3.80   & \multicolumn{1}{l}{CC-BY-4.0} \\
    \textbf{EmoTa}~\citep{emota} & ta    & SpkR, GR, ER & 2.90   & 1.06  & 9.79  & EACL \\
    \textbf{SFDUSC}~\citep{sfdusc_ph} & tl    & SpkR, GR, ASR & 4.86  & 2.03  & 21.49 & \multicolumn{1}{l}{CC BY-NC-ND 4.0} \\
    \textbf{YODAS2$^\dagger$}~\citep{yodas} & en, id, zh, th, vi & TLoc, TCQ, SQA & 493.12$^\dagger$ & 20.01 & 180.00   & \multicolumn{1}{l}{CC-BY 3.0} \\
    \textbf{ESD}~\citep{esd}   & en, zh & ASR, ER, SpkR & 6.56  & 1.41  & 10.79 &  MIT \\
    \textbf{M3ED}~\citep{m3ed} & zh    & ER, GR & 0.83  & 0.12  & 7.04  & \multicolumn{1}{l}{CC BY-NC-ND 4.0} \\
    \textbf{MIG}~\citep{mig} & my    & ASR, SpkR & 1.50   & 0.80   & 21.39 & \multicolumn{1}{l}{NA} \\
    \textbf{TEC}~\citep{tec} & ta    & ER    & 0.68  & 2.69  & 29.86 & NA\\
    \textbf{ASR-SgpCCSC$^\dagger$}~\citep{sgpccsc} & zh    & ASR, TLoc, TCQ, SQA   & 43.76$^\dagger$    & 0.88  & 180.00    & \multicolumn{1}{l}{CC BY-NC-ND 4.0} \\
    \bottomrule
    \end{tabular}
    }
  \vspace{2pt}
  
\begin{minipage}{0.98\textwidth}
\footnotesize
\textit{Duration breakdown.}
For long-form sources marked with $^\dagger$, the total hours are consist as follows:
\end{minipage}

\vspace{2pt}
\centering
\scriptsize
\begin{tabular}{lcc}
\toprule
Source & Short-form hours & Long-form hours \\
\midrule
SG Streets & 5.73 & 0.82 \\
YODAS2 & 26.77 & 466.35 \\
ASR-SgpCCSC & 31.89 & 11.87 \\
\bottomrule
\end{tabular}
\caption{\textcolor{revisionblue}{
Overview of source datasets, including language coverage, task types, selected audio duration, and licensing information.
Total duration includes both short-form ($\leq$30s) and long-form ($>$30s) evaluation instances. $^\dagger$ Source contributes long-form evaluation instances ($>$30s), the short-/long-form breakdown is reported below the table.}
}
\label{tab:catalog}%
\end{table*}%

\begin{table*}[!htbp]
\centering
\scriptsize
\setlength{\tabcolsep}{3pt}
\begin{tabular}{lcccccc}
\toprule
Dataset & Transcript & Time alignment & Speaker ID & Gender & Age & Emotion \\
\midrule
Common Voice          & \metaused     & \metana      & \metaused   & \metaused   & \metaused     & \metana      \\
FLEURS                & \metaused     & \metana      & \metana     & \metaused   & \metana       & \metana      \\
OpenSLR               & \metaused     & \metana      & \metana     & \metaused   & \metana       & \metana      \\
Bloom-Speech          & \metaused     & \metana      & \metana     & \metana     & \metana       & \metana      \\
Thai Elderly Speech   & \metaused     & \metana      & \metaused   & \metaused   & \metaunused   & \metana      \\
LOTUS                 & \metaused     & \metana      & \metana     & \metana     & \metana       & \metana      \\
THAI-SER              & \metaunused   & \metana      & \metaused   & \metaused   & \metaunused   & \metaused    \\
SG Streets            & \metaused     & \metaused    & \metana     & \metaused   & \metana       & \metana      \\
VietMed               & \metaused     & \metana      & \metana     & \metana     & \metana       & \metana      \\
VoxVietnam-O          & \metana       & \metana      & \metaused   & \metana     & \metana       & \metana      \\
Bud500                & \metaused     & \metana      & \metana     & \metana     & \metana       & \metana      \\
Vietnam-Celeb         & \metana       & \metana      & \metaunused & \metaused   & \metana       & \metana      \\
ASR-SMalDuSC          & \metaused     & \metana      & \metaused   & \metaused   & \metaunused   & \metana      \\
ASR-MALCSC            & \metaused     & \metana      & \metaunused & \metaunused & \metaunused   & \metana      \\
IndoWaveSentiment     & \metaunused   & \metana      & \metaunused & \metaused   & \metana       & \metaused    \\
EmoTa                 & \metaunused   & \metana      & \metaused   & \metaused   & \metaunused   & \metaused    \\
SFDUSC                & \metaused     & \metana      & \metaused   & \metaused   & \metana       & \metana      \\
YODAS2                & \metafiltered & \metaused    & \metana     & \metana     & \metana       & \metana      \\
ESD                   & \metaused     & \metana      & \metaused   & \metaunused & \metana       & \metaused    \\
M3ED                  & \metaunused   & \metana      & \metana     & \metaused   & \metaunused   & \metaused    \\
MIG                   & \metaused     & \metana      & \metaused   & \metaunused & \metaunused   & \metana      \\
TEC                   & \metana       & \metana      & \metana     & \metana     & \metana       & \metaused    \\
ASR-SgpCCSC           & \metaused     & \metaused    & \metana     & \metana     & \metana       & \metana      \\
\bottomrule
\end{tabular}

\vspace{2pt}
\begin{minipage}{0.85\textwidth}
\footnotesize
\textit{Notes.}
\metaused indicates source-provided metadata used in \textsc{SEA-Speechbench} for evaluation labels or references, filtering, pairing, timestamp construction, or split enforcement.
\metafiltered indicates that transcripts are used only after filtering or alignment validation.
\metaunused indicates that source-provided metadata are available but not used, due to issues such as severe class imbalance or unverified transcripts.
\metana \ indicates that metadata are not provided.
\end{minipage}
\caption{Metadata availability and usage for the source datasets listed in Table~\ref{tab:catalog}.}
\label{tab:metadata-availability}
\end{table*}

\section{Data Processing and Synthesis}~\label{app:processing}
We provide additional details on data processing and the construction of new datasets from source materials.

\subsection{Data Processing}
\textbf{Audio and Prompt Standardization:} All audio was resampled to 16 kHz for consistency and model compatibility. We provide parallel prompts in native SEA languages and English to reflect realistic usage patterns, constructing standardized templates where needed. Prompt examples are provided in the Appendix~\ref{app:prompt}.

\noindent\textbf{Dataset Sampling:} 1. We adopted official test splits when available or sampled 1,000 instances while preserving identifiers to prevent contamination. 2. For our evaluation set, we further sampled 1,000 instances per dataset using class-balanced sampling for classification tasks, ensuring computational feasibility while maintaining representativeness. 3. We checked that uniquely identifying information such as names were removed when curating and processing data from open-source materials. We also applied profanity and content filtering on raw speech transcripts in the YODAS2 dataset.

This standardized processing transforms diverse datasets into a unified, reproducible benchmark supporting consistent evaluation across models. 

{\color{revisionblue}
\paragraph{ER Label Normalization:}
\label{app:er-label-map}

SEA-SpeechBench unifies emotion labels across source datasets into a closed nine-class label set: \textit{neutral}, \textit{happy}, \textit{angry}, \textit{sad}, \textit{surprised}, \textit{disgusted}, \textit{fearful}, \textit{frustrated}, and \textit{disappointed}.
We obtain this set as the union of annotated emotion categories across the source datasets retained for ER, preserving fine-grained distinctions rather than collapsing them into a coarser taxonomy in order to retain the emotional nuances captured in the original annotations.

Normalization proceeds in two stages. First, each source dataset is mapped from its native label form to an intermediate English label set during dataset construction (e.g., Mandarin labels in ESD are mapped to their English equivalents; numeric codes in IndoWaveSentiment, TEC, and THAI-SER are mapped to lexical labels; three-letter codes in EmoTa are expanded to full words). Second, before the final dataset save, we apply a uniform lexical normalization step that lowercases all labels and converts each to a canonical adjectival form, so that variants such as \textit{Surprise} / \textit{surprise}, \textit{fear}, and \textit{disgust} are unified to \textit{surprised}, \textit{fearful}, and \textit{disgusted}, respectively. Table~\ref{tab:er-label-map} reports the source-to-canonical mapping after both stages. Source samples with missing or empty emotion labels were filtered out during construction; all retained samples were mapped deterministically.

\begin{table*}[!htbp]
\centering
\small
\renewcommand{\arraystretch}{1.2}
\resizebox{\linewidth}{!}{%
\begin{tabular}{lll}
\toprule
Source dataset & Source label form & Canonical labels in \textsc{SEA-SpeechBench} \\
\midrule
ESD (en, zh)            & \{Neutral, Angry, Happy, Sad, Surprise\} (zh/en mixed) & neutral, angry, happy, sad, surprised \\
EmoTa (ta)              & \{neu, sad, fea, ang, hap\}                            & neutral, sad, fearful, angry, happy \\
IndoWaveSentiment (id)  & \{01, 02, 03, 04, 05\}                                  & neutral, happy, surprised, disgusted, disappointed \\
M3ED (zh)               & \texttt{final\_main\_emo} (free-text English)           & neutral, angry, happy, sad, surprised, disgusted, fearful \\
TEC (ta)                & \{0, 1, 2, 3, 4\}                                       & neutral, angry, happy, sad, fearful \\
THAI-SER (th)           & \{0, 1, 2, 3, 4\}                                       & neutral, angry, happy, sad, frustrated \\
\midrule
\multicolumn{2}{l}{\textbf{Union (closed 9-class canonical set)}} & \multicolumn{1}{l}{\begin{tabular}[t]{@{}l@{}}neutral, happy, angry, sad, surprised,\\ disgusted, fearful, frustrated, disappointed\end{tabular}} \\
\bottomrule
\end{tabular}
}
\caption{ER source-to-canonical label mapping. All source labels are first mapped to an intermediate English label set during dataset construction, then uniformly lowercased and converted to a canonical adjectival form before the final dataset save (e.g., \textit{Surprise}$\to$\textit{surprised}, \textit{fear}$\to$\textit{fearful}, \textit{disgust}$\to$\textit{disgusted}). The union of canonical labels across all sources forms a closed 9-class set.}
\label{tab:er-label-map}
\end{table*}

\subsection{Label Balance Statistics}
\label{app:label-balance}

We report class-balance statistics for the four classification tasks in \textsc{SEA-Speechbench} (AgeR, ER, GR, SpkR). Statistics are computed at the per-dataset level, since per-dataset scores are first computed and then averaged across datasets. For each constituent dataset, we compute the normalized Shannon entropy
\begin{equation}
\tilde{H} = \frac{H}{\log K} \in [0, 1],
\end{equation}
where $H$ is the empirical label entropy, $K$ is the number of classes, and $\tilde{H} = 1$ corresponds to a perfectly uniform distribution. Table~\ref{tab:label-balance} reports the across-dataset mean and range of $\tilde{H}$ for each task.

\begin{table}[h]
\centering
\small
\begin{tabular}{lccc}
\toprule
Task & \#Classes & $\tilde{H}$ (mean) & $\tilde{H}$ (range) \\
\midrule
ER   & up to 7 & 0.978 & [0.943, 1.000] \\
GR   & 2       & 0.958 & [0.949, 1.000] \\
SpkR & 2       & 0.999 & [0.990, 1.000] \\
AgeR & up to 3 & 0.745 & [0.502, 0.912] \\
\bottomrule
\end{tabular}
\caption{Per-dataset normalized label entropy $\tilde{H}$ for the four classification tasks in \textsc{SEA-Speechbench}. Higher $\tilde{H}$ indicates a more uniform class distribution; $\tilde{H} = 1$ corresponds to a perfectly balanced label set.}
\label{tab:label-balance}
\end{table}

GR, SpkR, and ER exhibit near-uniform per-dataset distributions. AgeR is the sole exception ($\tilde{H} = 0.745$), reflecting demographic skew and digital participation patterns in the source data: Mozilla Common Voice age labels are self-reported by crowdsourced contributors, with younger and middle-aged adults over-represented relative to elderly speakers. We retain the source distribution rather than enforcing a uniform 3-way split, since forced rebalancing would discard a substantial portion of usable evaluation data.

To prevent the residual class imbalance from biasing evaluation, we adopt \textbf{macro-F1} as the primary metric for GR, SpkR, and AgeR (\S\ref{sec:metric}), which weights all classes equally regardless of support. For ER, the near-uniform per-dataset label distribution ($\tilde{H} = 0.978$) makes accuracy and macro-F1 nearly equivalent in expectation, so the choice between them is not driven by class balance. The harder issue is that emotion descriptions are frequently graded (e.g., \textit{``slightly frustrated''}) or compound (e.g., \textit{``happy and excited''}), making single-label string canonicalization noise-prone. We therefore retain \textbf{judge-based accuracy} for ER, motivated by the difficulty of reliable string matching rather than by metric convenience.
}

\subsection{New Evaluation Data Construction}~\label{app:synthesis}
\noindent
\textbf{SpkR (Speaker Recognition)}: We constructed datasets by sampling and concatenating pairs of audio segments, separated by a short beep, drawn either from the same speaker or from different speakers, to create controlled instances for the speaker verification task. Refer to Table~\ref{tab:catalog} for the datasets used to synthesize the speaker recognition task.

\textbf{SQA (Spoken Question Answering)}: We carefully filtered the YODAS2 dataset \citep{yodas} using CTC forced alignment scores between the audio clips and the provided transcriptions with a log-probability threshold of -1, exact language match between the audio and text transcription, as well as profanity and content filtering. Based on the cleaned transcriptions, we generated question–answer pairs using GPT-4.1~\citep{gpt4o}.

To ensure the quality of generated SQA instances, we further conducted a post-generation human audit. For each audited batch, a native-speaker annotator reviewed 50 randomly sampled generated question--answer pairs along three axes: factual grounding in the source audio, answer correctness, and question-type diversity. Items failing any of these criteria were corrected or removed, and recurring failure modes were used to revise the generation prompt. We repeated this audit-and-revision loop until at least $90\%$ of audited items passed all three criteria. The final accepted SQA batch was then generated using the converged prompt.

To examine the robustness of our results to potential generator--judge--evaluatee overlap, we re-scored the SQA outputs generated by GPT-4o and the other evaluated models using Gemma-3-27B-Instruct, and compared the resulting model rankings with those obtained using GPT-4.1. The rankings exhibit high agreement across judges, with Spearman $\rho = 0.967$ for English prompts and $\rho = 0.885$ for SEA prompts as shown in Figure~\ref{tab:judge_robustness}. The overall ranking and main SQA conclusions are therefore robust to the choice of judge. In particular, GPT-4o ranks second under GPT-4.1 judging and fourth under Gemma judging for English and SEA prompt languages, suggesting that its exact rank is somewhat sensitive to the choice of judge, while the broader conclusions remain unchanged.

\begin{table}[t]
    \centering
    \resizebox{\columnwidth}{!}{%
    \begin{tabular}{lccc}
        \toprule
        Prompt language & Spearman $\rho$ & \multicolumn{2}{c}{GPT-4o Rank} \\
        \cmidrule(lr){3-4}
        & & GPT-4.1 judge & Gemma judge \\
        \midrule
        English      & 0.967 & 2 / 13 & 4 / 13 \\
        SEA          & 0.885 & 2 / 13 & 4 / 13 \\
        Avg. ENG+SEA & 0.951 & 2 / 13 & 4 / 13 \\
        \bottomrule
    \end{tabular}%
    }
    \caption{Correlation between model rankings for SQA under GPT-4.1 and Gemma-3-27B-Instruct judging, along with the rank of GPT-4o under each judge.}
    \label{tab:judge_robustness}
\end{table}

\textbf{TCQ (Timestamped Content Query)}: We filtered the YODAS2 dataset following the same procedure used for SQA. We then identified utterances whose CTC log-probability exceeded -0.1, recording their start and end timestamps. The task was formulated such that models are required to transcribe utterances occurring between the specified timestamps.

\textbf{TLoc (Temporal Localization)}: We adopted the same preprocessing procedure as in TCQ. The task was formulated such that models are required to predict the start and end timestamps corresponding to specified utterances.

In all cases, we applied the same standardized audio preprocessing, sample selection, and prompt construction pipeline, ensuring these synthesized datasets are consistent and compatible with the broader benchmark.

\section{Definition of Prompt Advantage Score (PAS)}~\label{app:pas}
We quantify prompt-language effects with a scale-invariant score capturing both \emph{magnitude} and \emph{direction}. Let $s_{\text{SEA}}$ and $s_{\text{EN}}$ denote task scores for SEA language prompt and English prompt, respectively. For each model $M$, task $T$, and language $L$, define the symmetric relative difference
\begin{equation}
d_{M,T,L}
=
\frac{\bigl|\,s_{\text{SEA}}-s_{\text{EN}}\,\bigr|}
{\tfrac{|s_{\text{SEA}}|+|s_{\text{EN}}|}{2}+\varepsilon},
\end{equation}
and aggregate across models within task by the median
\(\tilde d_{T,L}=\operatorname{median}\nolimits_{M} d_{M,T,L}\).
Directionality is encoded by the model-wise win rate of local prompts
\(w_{T,L}=\mathbb{P}_{M}\!\bigl(s_{\text{SEA}}>s_{\text{EN}}\bigr)\).
The task-level Prompt Advantage Score combines direction and effect size:
\begin{equation}
\mathrm{PAS}_{T,L}=\bigl(2(w_{T,L}-0.5)\bigr)\cdot \tilde d_{T,L},
\end{equation}
so $\mathrm{PAS}_{T,L}>0$ favors local prompts and $\mathrm{PAS}_{T,L}<0$ favors English. The language-level score averages over the tasks covered by $L$, denoted $\mathcal{T}(L)$:
\begin{equation}
\mathrm{PAS}_{L}=\frac{1}{|\mathcal{T}(L)|}\sum_{T\in \mathcal{T}(L)} \mathrm{PAS}_{T,L}.
\end{equation}
By construction, $|\mathrm{PAS}_{L}|$ measures the strength of prompt sensitivity while $\operatorname{sign}(\mathrm{PAS}_{L})$ indicates the preferred prompt language. 

For tasks where lower scores indicate better performance 
(ASR and TCQ, measured by WER/CER), we first transform 
the raw scores via $s \leftarrow \frac{1}{1 + s}$ before 
computing $d_{M,T,L}$ and $w_{T,L}$, mapping the metric 
into a higher-is-better scale on $(0, 1]$. This ensures 
that $s_{\text{SEA}} > s_{\text{EN}}$ consistently 
corresponds to SEA prompts yielding better performance, 
and $\text{PAS}_{T,L} > 0$ uniformly indicates local 
prompt advantage regardless of the original metric 
direction.

\section{Text-based Language Understanding}
Prompt-language performance differences in speech--text models cannot be fully explained by their text-based language understanding. To separate text-only cross-linguistic effects from cross-modal behavior, we conducted a prompt-only sanity check on the Age Recognition templates in ta, th, vi, and zh. Specifically, we ask each model (i) to translate the prompt into English and (ii) to state in one sentence what output is required; the results are shown in Table~\ref{tab:age_recognition_gap}. For MERaLiON-2-3B, it answers 8/8 correctly for both requests across ta/th/vi/zh, confirming that prompt comprehension is preserved in isolation. Despite this perfect text-only comprehension, the model exhibits a consistent English advantage on Age Recognition across all four languages, with the gap ranging from $+$1.64 (Thai) to $+$15.23 (Tamil). This pattern indicates that the observed prompt asymmetry cannot be attributed solely to discrepancies in language understanding, and instead points to a broader cross-modal instruction-following or alignment gap under non-English prompts.

\begin{table}[!htbp]
\centering
\vspace{-0.3cm}
\resizebox{\linewidth}{!}{
\begin{tabular}{lcc}
\toprule
Prompt Language & $\Delta$ Score (English $-$ Native) & Prompt Check Failure Rate \\
\midrule
Tamil (ta)      & $+$15.23 & 0 \\
Thai (th)       & $+$1.64  & 0 \\
Vietnamese (vi) & $+$5.64  & 0 \\
Chinese (zh)    & $+$2.38  & 0 \\
\bottomrule
\end{tabular}
}
\caption{Performance gap on the AgeR task between English and SEA language prompts for MERaLiON-2-3B. A positive $\Delta$ indicates that English prompts outperform native-language prompts. Prompt Check Failure Rate measures the fraction of prompts where the model failed a text-only comprehension check.}
\label{tab:age_recognition_gap}
\vspace{-0.3cm}
\end{table}

\section{Postprocessing of Evaluation}
To ensure fair and consistent evaluation across Southeast Asian languages, we design a unified text normalization pipeline comprising three sequential stages:

\noindent\textbf{Canonical Form Conversion.} Transcripts undergo initial standardization through NFC Unicode normalization and systematic lowercasing to eliminate encoding inconsistencies and case variations. Configurable punctuation filtering preserves linguistically meaningful characters (hyphens, apostrophes) while removing extraneous symbols that could introduce evaluation noise.

\noindent\textbf{Content Filtering and Lexical Standardization.} Non-linguistic artifacts including speaker tags, bracketed annotations, and conversational fillers are systematically removed. Digital content undergoes targeted normalization, and English contractions receive consistent expansion. For character-based writing systems, e.g., Chinese and Thai, inter-character spacing is inserted to enable reliable tokenization.

\noindent\textbf{Language-Specific Enhancement and Validation.} We collaborated with native speakers across all SEA languages to validate and refine normalization heuristics.

\section{Evaluation Prompt Examples}~\label{app:prompt}
\begin{figure*}[!ht]
  \centering
  \caption{Overview of prompt examples: illustrative prompts provided for ASR, ST, SQA, AgeR, and ER tasks across multiple languages.}
  \includegraphics[width=0.99\linewidth]{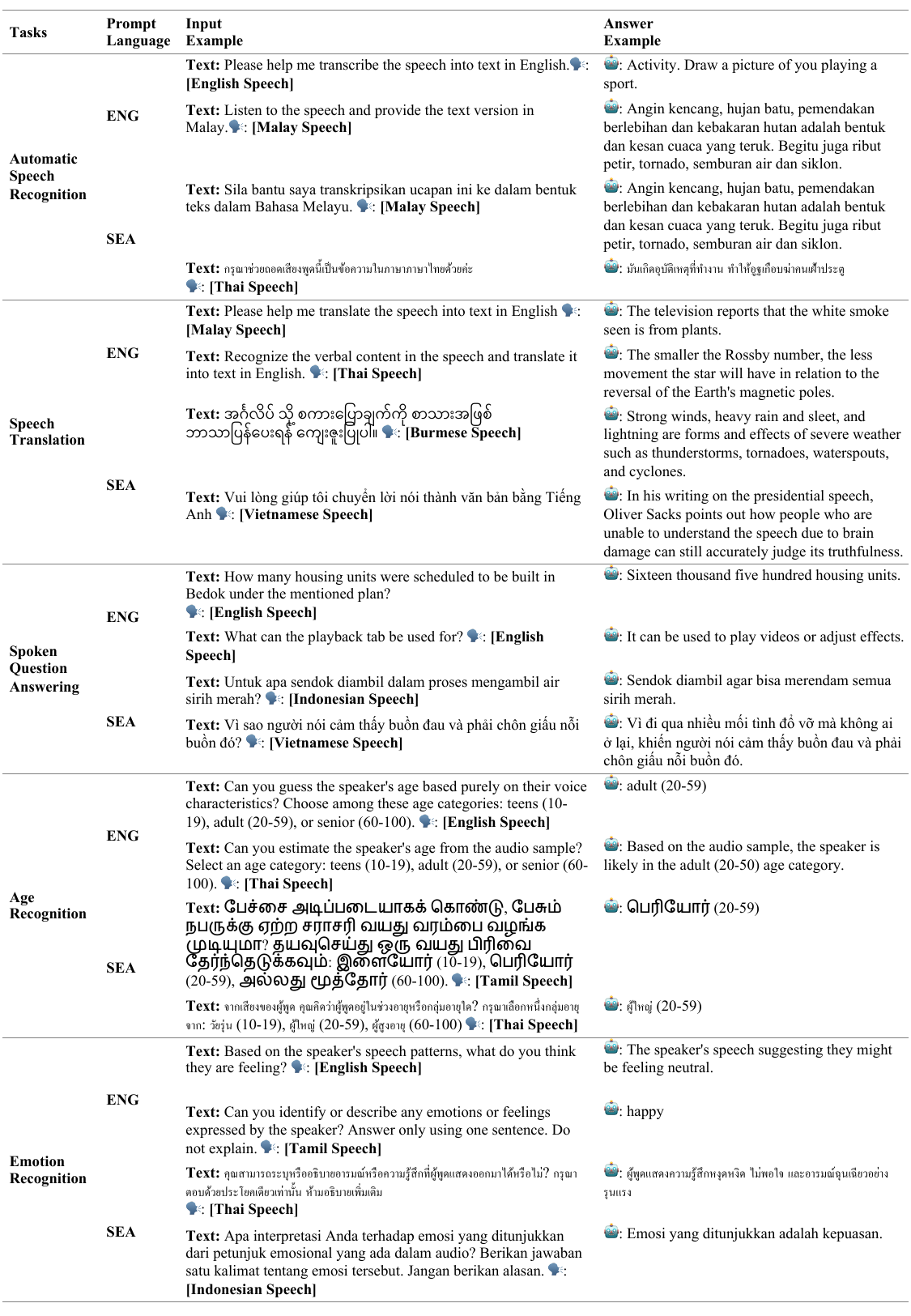}
  \label{fig:task_prompt_example_a}
\end{figure*}

\begin{figure*}[!ht]
  \centering
  \caption{Overview of prompt examples: illustrative prompts provided for GR, SpkR, TCQ, and TLoc tasks across multiple languages.}
  \includegraphics[width=0.99\linewidth]{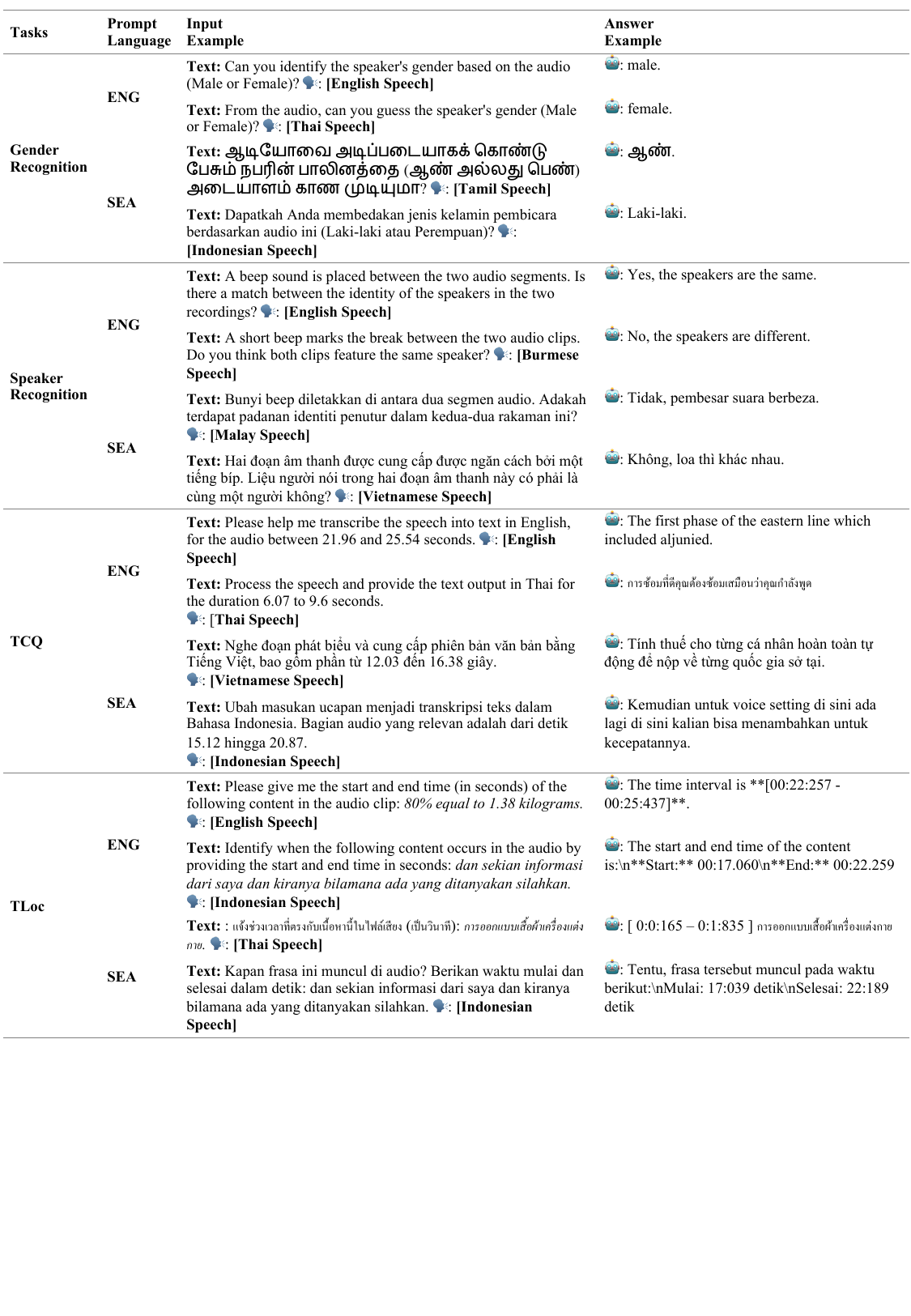}
  \label{fig:task_prompt_example_b}
\end{figure*}

We evaluate using parallel prompts in native SEA languages and English. Prompt examples are provided in Figure~\ref{fig:task_prompt_example_a} and \ref{fig:task_prompt_example_b}.

\clearpage

\clearpage
\section{Model-as-Judge Prompt Examples}~\label{app:judge_prompt}

\begin{table}[!ht]
  \centering
  \fontsize{8.5pt}{10pt}\selectfont 
\caption{Model-as-Judge prompts, adopted from \protect\cite{audiobench}. To ensure protocol consistency, human judges followed the same evaluation instruction.}
    \begin{tabularx}{\textwidth}{l | X }
    \toprule
    \textbf{Task} & \textbf{Prompt} \\
    \midrule
    \textbf{SQA} & 
    \it [Question]\newline
    \{question\}\newline
    \par
    [Reference]\newline
    \{reference\}\newline
    \par
    [Model Prediction]\newline
    \{prediction\}\newline
    \par
    [Task]\newline
    Rate the model prediction based on its alignment with the reference, focusing on accuracy and relevance to the reference. Be critical.\newline
    Score0: The prediction repeats or rephrases the question without giving an answer.\newline
    Score0: The prediction is refusing to give concrete results, providing something like 'cannot decide'.\newline
    Score0: The prediction is completely misaligned, providing incorrect or irrelevant information compared to the reference. \newline
    Score1: The prediction shows minimal alignment, often misunderstanding or providing irrelevant details unrelated to the reference. \newline
    Score2: The prediction recognizes the topic but diverges significantly from the reference in accuracy or relevance. \newline
    Score3: The prediction aligns with the reference generally but lacks detail or precise accuracy in some aspects. \newline
    Score4: The prediction is mostly accurate and relevant, closely following the reference but could be clearer or more detailed. \newline
    Score5: The prediction is highly accurate, detailed, and matches the reference perfectly, capturing its essence and detail. \newline
    \par
    Your response should be formatted as follows:
    Explanation: (Provide a concise explanation of your rating, comparing the reference with the model prediction. "The reference is [XXX], while the model prediction is [YYY]. I think ...")
    Rating: (int)\\
    \midrule
    \bf \makecell{AgeR \\ ER \\ GR \\ SpkR} & 
    \it [Question]\newline
    \{question\}\newline
    \par
    [Reference]\newline
    \{reference\}\newline
    \par
    [Model Prediction]\newline
    \{prediction\}\newline
    \par
    [Task]\newline
    Rate the model prediction based on its alignment with the reference, focusing on accuracy and relevance to the reference. Be critical.\newline
    Score0: The prediction repeats or rephrases the question without giving an answer.\newline
    Score0: The prediction is refusing to give concrete results, providing something like 'cannot decide'.\newline
    Score0: The prediction is wrong, providing incorrect or irrelevant information compared to the reference.\newline
    Score1: The prediction is correct, capturing or covering the meaning from the reference.\newline
    \par
    Your response should be formatted as follows:
    Explanation: (Provide a concise explanation of your rating, comparing the reference with the model prediction. "The reference is [XXX], while the model prediction is [YYY]. I think ...")
    Rating: (int)\\
    \bottomrule
    \end{tabularx}
  \label{tab:judge_prompt}
\end{table}

\clearpage
\section{Detailed Evaluation Results}\label{app:evaluation_detail}

\begin{table*}[!ht]
\centering
\caption{ASR performance: results across datasets comparing English and SEA prompts. Scores are reported as raw WER and CER values without  normalization; lower values indicate better performance.}
\label{tab:my-table}
\resizebox{0.95\linewidth}{!}{%
\renewcommand{\arraystretch}{1.2}
\begin{tabular}{
    c@{\;}c@{\;}c@{\;}c@{\;}|
    *{13}{c@{\hspace{0.3em}}}
    c
}
\toprule
\multicolumn{4}{c|}{\textbf{Model}}
    & \makecell{\textbf{MERaLiON2} \\ \textbf{10B}} 
    & \makecell{\textbf{MERaLiON2} \\ \textbf{3B}} 
    & \makecell{\textbf{SeaLLMs} \\ \textbf{Audio} \\ \textbf{7B}}
    & \makecell{\textbf{Phi-4} \\ \textbf{multi-} \\ \textbf{-modal} \\ \textbf{instruct}}
    & \makecell{\textbf{Kimi} \\ \textbf{Audio}} 
    & \makecell{\textbf{Voxtral} \\ \textbf{mini}} 
    & \makecell{\textbf{Qwen2} \\ \textbf{Audio} \\ \textbf{7B} \\ \textbf{Instruct}} 
    & \makecell{\textbf{Qwen} \\ \textbf{2.5} \\ \textbf{Omni} \\ \textbf{3B}}
    & \makecell{\textbf{Qwen} \\ \textbf{2.5} \\ \textbf{Omni} \\ \textbf{7B}}
    & \makecell{\textbf{gemma} \\ \textbf{3n} \\ \textbf{E4B-it}}
    & \makecell{\textbf{gemma} \\ \textbf{3n} \\ \textbf{E2B-it}}
    & \makecell{\textbf{Gemini} \\ \textbf{2.5} \\ \textbf{Flash}}
    & \makecell{\textbf{Whisper} \\ \textbf{large} \\ \textbf{v3}}
    & \makecell{\textbf{GPT} \\ \textbf{4o} \\ \textbf{Audio}} \\ 
  \midrule
\multicolumn{4}{c|}{size} & 10B  & 3B   & 7B   & 5.6B & 7B    & 3B   & 7B   & 3B    & 7B   & 4B    & 2B   & -    & 1.5B          & -             \\ \midrule
\textbf{Data} & \textbf{Metrics} & \textbf{Lang} & \textbf{Prompt} &      &      &      &      &       &      &      &       &      &       &      &      & \multicolumn{2}{c}{No Prompt} \\ \cmidrule{1-4} \cmidrule{17-18} 
\bf Bloom-Speech  & WER    & en   & ENG    & 0.06 & 0.07 & 0.74 & 0.06 & 0.06  & 0.20 & 0.13 & 0.06  & 0.06 & 0.38  & 0.52 & 0.05 & 0.04          & 0.06          \\
              &        &      & SEA    & 0.06 & 0.07 & 0.74 & 0.06 & 0.06  & 0.20 & 0.13 & 0.06  & 0.06 & 0.38  & 0.52 & 0.05 & -          & -             \\
              & CER    & my   & ENG    & 0.74 & 0.99 & 1.08 & 2.29 & 2.50  & 2.94 & 0.98 & 2.74  & 4.19 & 2.36  & 4.94 & 0.73 & 1.55          & 0.98          \\
              &        &      & SEA    & 0.78 & 0.98 & 1.19 & 4.82 & 9.57  & 3.33 & 1.72 & 1.30  & 3.24 & 6.56  & 8.36 & 0.75 & -             & -             \\
              & WER    & tl   & ENG    & 0.14 & 0.14 & 0.84 & 5.60 & 9.00  & 3.35 & 1.19 & 0.41  & 0.50 & 0.23  & 0.63 & 0.09 & 0.12          & 0.11          \\
              &        &      & SEA    & 0.17 & 0.17 & 1.37 & 3.10 & 4.06  & 2.24 & 3.10 & 0.41  & 1.84 & 0.58  & 1.48 & 0.08 & -             & -             \\
\bf Bud500        & WER    & vi   & ENG    & 0.10 & 0.05 & 0.11 & 1.50 & 38.71 & 1.62 & 1.45 & 0.08  & 0.07 & 7.09  & 8.89 & 0.15 & 0.47          & 0.21          \\
              &        &      & SEA    & 0.11 & 0.34 & 0.11 & 3.55 & 28.94 & 3.48 & 2.89 & 0.08  & 0.19 & 6.68  & 7.18 & 0.15 & -             & -             \\
\bf Commonvoice          & WER    & en   & ENG    & 0.08 & 0.09 & 0.13 & 0.08 & 0.07  & 0.10 & 0.13 & 0.08  & 0.07 & 0.23  & 0.38 & 0.11 & 0.09          & 0.10          \\
              &        &      & SEA    & 0.08 & 0.09 & 0.13 & 0.08 & 0.07  & 0.10 & 0.13 & 0.08  & 0.07 & 0.23  & 0.38 & 0.11    & -             & -             \\
              & WER    & id   & ENG    & 0.11 & 0.11 & 0.10 & 1.44 & 0.68  & 0.38 & 0.74 & 0.11  & 0.10 & 0.47  & 0.25 & 0.03 & 0.08          & 0.06          \\
              &        &      & SEA    & 0.07 & 0.13 & 0.19 & 1.61 & 0.60  & 0.37 & 0.79 & 0.12  & 0.10 & 0.68  & 1.40 & 0.03 & -             & -             \\
              & WER    & ta   & ENG    & 0.50 & 0.52 & 1.49 & 1.83 & 4.34  & 1.33 & 1.41 & 1.13  & 1.41 & 0.68  & 1.95 & 0.29 & 0.64          & 0.46          \\
              &        &      & SEA    & 0.33 & 0.58 & 1.41 & 3.36 & 2.34  & 1.37 & 1.97 & 1.69  & 1.13 & 2.33  & 2.93 & 0.28 & -             & -             \\
              & CER    & th   & ENG    & 0.16 & 0.1  & 0.05 & 2.35 & 1.38  & 0.53 & 1.31 & 0.14  & 0.76 & 0.51  & 4.30 & 0.03 & 0.07          & 0.07          \\
              &        &      & SEA    & 0.09 & 0.21 & 0.05 & 8.41 & 1.14  & 0.45 & 1.27 & 0.16  & 0.78 & 11.41 & 5.79 & 0.04 & -             & -             \\
              & WER    & vi   & ENG    & 0.46 & 0.56 & 0.48 & 1.25 & 0.86  & 0.78 & 1.34 & 0.49  & 0.48 & 1.10  & 1.07 & 0.09 & 0.45          & 0.14          \\
              &        &      & SEA    & 0.16 & 0.73 & 0.48 & 1.72 & 1.02  & 0.76 & 0.11 & 0.48  & 0.47 & 1.28  & 0.97 & 0.09 & -             & -             \\
              & CER    & zh   & ENG    & 0.12 & 0.14 & 0.10 & 0.13 & 0.07  & 0.47 & 0.24 & 0.06  & 0.05 & 0.80  & 2.62 & 0.24 & 0.18          & 0.11          \\
              &        &      & SEA    & 0.12 & 0.16 & 0.08 & 0.07 & 0.05  & 0.44 & 0.12 & 0.06  & 0.05 & 2.14  & 2.36 & 0.11 & -             & -             \\
\bf ESD           & WER    & en   & ENG    & 0.05 & 0.05 & 0.06 & 0.03 & 0.05  & 0.16 & 0.08 & 0.04  & 0.04 & 0.14  & 0.27 & 0.05 & 0.04          & 0.06          \\
              &        &      & SEA    & 0.05 & 0.05 & 0.06 & 0.03 & 0.05  & 0.16 & 0.08 & 0.04  & 0.04 & 0.14  & 0.27 & 0.05    & -             & -             \\
              & CER    & zh   & ENG    & 0.05 & 0.05 & 0.11 & 0.05 & 0.02  & 0.33 & 0.21 & 0.02  & 0.02 & 0.51  & 1.76 & 0.12 & 0.04          & 0.05          \\
              &        &      & SEA    & 0.05 & 0.05 & 0.07 & 0.02 & 0.02  & 0.39 & 0.05 & 0.02  & 0.02 & 0.74  & 1.11 & 0.08 & -             & -             \\
\bf FLEURS        & WER    & tl   & ENG    & 0.16 & 0.18 & 1.06 & 5.92 & 0.94  & 1.41 & 1.98 & 0.80  & 0.57 & 0.07  & 0.16 & 0.07 & 0.12          & 0.07          \\
              &        &      & SEA    & 0.16 & 0.22 & 1.15 & 4.34 & 0.60  & 1.01 & 1.65 & 0.89  & 1.83 & 0.17  & 0.18 & 0.08 & -             & -             \\
              & WER    & id   & ENG    & 0.06 & 0.10 & 0.45 & 4.07 & 0.47  & 0.34 & 0.74 & 0.18  & 0.10 & 0.54  & 0.09 & 0.03 & 0.07          & 0.04          \\
              &        &      & SEA    & 0.06 & 0.12 & 0.12 & 3.19 & 0.36  & 0.28 & 0.64 & 0.11  & 0.31 & 0.10  & 0.09 & 0.03 & -             & -             \\
              & CER    & km   & ENG    & 0.86 & 1.71 & 1.00 & 4.42 & 8.21  & 1.14 & 1.06 & 3.23  & 4.55 & 1.12  & 4.90 & 0.17 & 1.35          & 0.31          \\
              &        &      & SEA    & 0.98 & 1.43 & 2.16 & 2.25 & 4.42  & 0.92 & 1.40 & 11.63 & 4.48 & 2.41  & 7.83 & 0.20 & -             & -             \\
              & CER    & lo   & ENG    & 0.39 & 0.86 & 1.01 & 2.45 & 4.52  & 1.29 & 1.08 & 1.39  & 2.63 & 0.15  & 1.42 & 0.16 & 1.02          & 0.38          \\
              &        &      & SEA    & 0.51 & 0.81 & 1.36 & 1.53 & 6.34  & 1.24 & 1.20 & 1.44  & 1.80 & 1.75  & 2.60 & 0.22 & -             & -             \\
              & WER    & ms   & ENG    & 0.11 & 0.14 & 0.76 & 3.34 & 1.32  & 0.44 & 1.03 & 0.28  & 0.27 & 0.86  & 0.18 & 0.05 & 0.08          & 0.06          \\
              &        &      & SEA    & 0.11 & 0.19 & 0.29 & 2.74 & 0.35  & 0.42 & 0.87 & 0.30  & 0.17 & 0.15  & 0.15 & 0.05 & -             & -             \\
              & CER    & my   & ENG    & 0.79 & 1.19 & 1.10 & 1.85 & 10.20 & 3.62 & 1.07 & 2.62  & 4.43 & 0.15  & 3.16 & 0.13 & 1.16          & 0.44          \\
              &        &      & SEA    & 0.76 & 1.03 & 1.10 & 4.35 & 8.87  & 3.07 & 1.11 & 1.99  & 4.01 & 1.42  & 1.97 & 0.12 & -             & -             \\
              & CER    & th   & ENG    & 0.14 & 0.13 & 0.10 & 4.14 & 2.93  & 0.57 & 1.04 & 0.16  & 1.20 & 0.15  & 4.03 & 0.07 & 0.09          & 0.04          \\
              &        &      & SEA    & 0.11 & 0.22 & 0.10 & 5.40 & 2.45  & 0.49 & 1.14 & 0.15  & 0.95 & 1.42  & 4.73 & 0.04 & -             & -             \\
              & WER    & vi   & ENG    & 0.10 & 0.16 & 0.91 & 4.46 & 2.39  & 0.40 & 1.05 & 0.10  & 0.09 & 0.17  & 0.30 & 0.06 & 0.08          & 0.04          \\
              &        &      & SEA    & 0.08 & 0.26 & 0.16 & 4.11 & 1.53  & 0.34 & 1.42 & 0.11  & 0.09 & 0.19  & 0.32 & 0.04 & -             & -             \\
              & CER    & zh   & ENG    & 0.10 & 0.12 & 0.84 & 0.1  & 0.08  & 0.36 & 0.18 & 0.07  & 0.06 & 0.25  & 1.00 & 0.18 & 0.08          & 0.07          \\
              &        &      & SEA    & 0.10 & 0.11 & 0.84 & 0.07 & 0.08  & 0.35 & 0.10 & 0.07  & 0.06 & 1.00  & 1.00 & 0.09 & -             & -             \\
\bf ASR-MALCSC        & WER    & ms   & ENG    & 0.21 & 0.26 & 0.79 & 3.82 & 14.23 & 1.30 & 1.77 & 0.21  & 1.01 & 4.86  & 2.81 & 0.26 & 0.31          & 0.42          \\
              &        &      & SEA    & 0.22 & 0.27 & 0.71 & 2.66 & 7.70  & 1.84 & 1.68 & 0.71  & 1.24 & 4.07  & 4.58 & 0.26 & -             & -             \\
\bf MIG           & CER    & my   & ENG    & 1.06 & 1.59 & 1.47 & 2.58 & 12.47 & 8.30 & 1.47 & 1.12  & 9.45 & 8.13  & 1.63 & 0.31 & 2.55          & 0.78          \\
              &        &      & SEA    & 0.97 & 1.53 & 2.09 & 9.35 & 15.58 & 2.80 & 2.09 & 1.64  & 4.43 & 3.58  & 5.53 & 0.29 & -             & -             \\
\bf OpenSLR       & CER    & km   & ENG    & 0.68 & 1.72 & 1.14 & 1.80 & 2.33  & 2.29 & 1.10 & 2.71  & 1.61 & 2.29  & 2.94 & 0.09 & 4.70          & 0.29          \\
              &        &      & SEA    & 0.88 & 1.74 & 2.18 & 5.91 & 4.64  & 2.80 & 1.13 & 8.42  & 1.70 & 2.89  & 3.92 & 0.30 & -             & -             \\
              & CER    & my   & ENG    & 0.71 & 0.97 & 1.16 & 1.80 & 6.78  & 5.34 & 1.16 & 1.57  & 3.88 & 1.26  & 1.51 & 0.05 & 2.51          & 0.42          \\
              &        &      & SEA    & 0.74 & 1.07 & 2.63 & 7.51 & 13.1  & 5.52 & 2.13 & 0.97  & 4.40 & 0.99  & 1.42 & 0.06 & -             & -             \\
              & WER    & ta   & ENG    & 0.26 & 0.34 & 1.55 & 2.03 & 4.58  & 1.06 & 1.55 & 1.58  & 1.26 & 0.22  & 0.45 & 0.24 & 2.51          & 0.35          \\
              &        &      & SEA    & 0.26 & 0.40 & 1.30 & 3.59 & 2.47  & 1.10 & 2.22 & 1.58  & 1.47 & 0.71  & 0.85 & 0.25 & -             & -             \\
\bf SFDUSC    & WER    & tl   & ENG    & 0.24 & 0.27 & 1.20 & 3.77 & 0.31  & 2.44 & 2.48 & 0.59  & 0.59 & 0.42  & 0.48 & 0.23 & 0.25          & 0.24          \\
              &        &      & SEA    & 0.25 & 0.10 & 1.11 & 3.82 & 1.30  & 2.24 & 4.07 & 0.70  & 0.61 & 0.39  & 0.75 & 0.21 & -             & -             \\
\bf ASR-SgpCCSC      & CER    & zh   & ENG    & 0.07 & 0.11 & 0.30 & 0.13 & 0.04  & 0.62 & 0.19 & 0.05  & 0.05 & 1.19  & 0.80 & 0.19 & 1.00          & 0.13          \\
              &        &      & SEA    & 0.07 & 0.09 & 0.21 & 3.82 & 0.04  & 0.71 & 0.21 & 0.05  & 0.04 & 5.55  & 1.77 & 0.12 & -             & -             \\
\bf SG Streets   & WER    & en   & ENG    & 0.10 & 0.05 & 0.57 & 0.17 & 0.82  & 0.39 & 0.20 & 0.09  & 0.09 & 2.27  & 4.11 & 0.22 & 1.00          & 0.29          \\
              &        &      & SEA    & 0.10 & 0.05 & 0.57 & 0.17 & 0.82  & 0.39 & 0.20 & 0.09  & 0.09 & 2.27  & 4.11 & 0.22 & -             & -             \\
\bf ASR-SMalDuSC      & WER    & ms   & ENG    & 0.07 & 0.11 & 0.27 & 2.28 & 2.21  & 0.55 & 0.27 & 0.21  & 0.22 & 0.23  & 0.18 & 0.02 & 1.00          & 0.03          \\
              &        &      & SEA    & 0.07 & 0.10 & 0.27 & 2.24 & 1.30  & 0.56 & 1.08 & 0.21  & 0.23 & 0.26  & 0.24 & 0.02 & -             & -             \\
\bf Thai Elderly Speech & CER    & th   & ENG    & 0.08 & 0.07 & 0.04 & 2.07 & 1.56  & 0.71 & 1.35 & 0.14  & 0.17 & 0.54  & 2.72 & 0.07 & 0.06          & 0.10          \\
              &        &      & SEA    & 0.08 & 0.22 & 0.04 & 8.38 & 1.16  & 0.75 & 1.32 & 0.20  & 0.12 & 7.46  & 2.63 & 0.05 & -             & -             \\
\bf LOTUS   & CER    & th   & ENG    & 0.02 & 0.03 & 0.02 & 3.35 & 4.51  & 0.46 & 1.14 & 0.04  & 0.03 & 0.06  & 1.06 & 0.01 & 0.03          & 0.01          \\
              &        &      & SEA    & 0.02 & 0.12 & 0.03 & 6.51 & 1.37  & 0.50 & 0.03 & 0.04  & 0.03 & 1.25  & 1.17 & 0.03 & -             & -             \\
\bf VietMed       & WER    & vi   & ENG    & 0.27 & 0.64 & 0.27 & 4.47 & 21.47 & 0.92 & 0.27 & 0.40  & 0.35 & 2.49  & 4.80 & 0.18 & 0.65          & 0.44          \\
              &        &      & SEA    & 0.28 & 0.49 & 0.27 & 3.23 & 19.44 & 1.48 & 1.56 & 0.40  & 0.26 & 2.71  & 4.07 & 0.19 & -             & -             \\ \midrule
\bf Average       &        &      & ENG    & 0.27 & 0.41 & 0.65 & 2.29 & 4.85  & 1.42 & 0.95 & 0.67  & 1.24 & 1.27  & 2.01 & 0.14 & 0.74          & 0.22          \\
              &        &      & SEA    & 0.27 & 0.43 & 0.74 & 3.39 & 4.30  & 1.45 & 1.20 & 1.11  & 1.08 & 2.28  & 2.50 & 0.14 & -             & -             \\ \bottomrule
\end{tabular}%
}
\end{table*}

\clearpage

\begin{table*}[!ht]
    \centering
    \caption{Age Recognition performance: results across datasets comparing English and SEA prompts. Evaluation is conducted using the macro-F1 metric ($\%$), where higher scores indicate better performance.}
    \label{tab:sea_results_ager}
    \resizebox{0.99\linewidth}{!}{%
    \renewcommand{\arraystretch}{1.5}
    \begin{tabular}{
        c@{\;}c@{\;}c@{\;}|
        *{12}{c@{\hspace{0.3em}}}
        c
        }
    \toprule
    \multicolumn{3}{c|}{\textbf{Model}}
    & \makecell{\textbf{MERaLiON2} \\ \textbf{10B}} 
    & \makecell{\textbf{MERaLiON2} \\ \textbf{3B}} 
    & \makecell{\textbf{SeaLLMs} \\ \textbf{Audio} \\ \textbf{7B}}
    & \makecell{\textbf{Phi-4} \\ \textbf{multi-} \\ \textbf{-modal} \\ \textbf{instruct}}
    & \makecell{\textbf{Kimi} \\ \textbf{Audio}} 
    & \makecell{\textbf{Voxtral} \\ \textbf{mini}} 
    & \makecell{\textbf{Qwen2} \\ \textbf{Audio} \\ \textbf{7B} \\ \textbf{Instruct}} 
    & \makecell{\textbf{Qwen} \\ \textbf{2.5} \\ \textbf{Omni} \\ \textbf{3B}}
    & \makecell{\textbf{Qwen} \\ \textbf{2.5} \\ \textbf{Omni} \\ \textbf{7B}}
    & \makecell{\textbf{gemma} \\ \textbf{3n} \\ \textbf{E4B-it}}
    & \makecell{\textbf{gemma} \\ \textbf{3n} \\ \textbf{E2B-it}}
    & \makecell{\textbf{Gemini} \\ \textbf{2.5} \\ \textbf{Flash}} 
    & \makecell{\textbf{GPT} \\ \textbf{4o} \\ \textbf{Audio}} \\ 
    \multicolumn{3}{c|}{Size} & 10B & 3B & 7B & 5.6B & 7B & 3B & 7B & 3B & 7B & 4B & 2B & - & - \\ 
    \midrule
    \textbf{Data} 
    & \textbf{Lang} 
    & \textbf{Prompt} 
    & & & & & & & & & & & & & \\ 
    \cmidrule(lr){1-3}
    \bf Commonvoice & en & ENG & 24.84 & 27.21 & 24.66 & 24.25 & 26.25 & 27.87 & 30.93 & 16.81 & 9.91 & 23.87 & 21.61 & 43.32 & 25.38 \\ 
    &  & SEA & 24.84 & 27.21 & 24.66 & 24.25 & 26.25 & 27.87 & 30.93 & 16.81 & 9.91 & 23.87 & 21.61 & 43.32 & 25.38 \\ 
    & ta & ENG & 28.57 & 30.28 & 18.56 & 27.20 & 43.13 & 34.69 & 16.76 & 31.93 & 29.42 & 26.79 & 22.93 & 33.33 & 27.31 \\ 
    &  & SEA & 20.99 & 15.05 & 1.62 & 14.66 & 12.40 & 18.62 & 9.13 & 7.58 & 6.62 & 20.58 & 15.65 & 41.00 & 32.82 \\ 
    & th & ENG & 38.64 & 40.16 & 39.53 & 29.57 & 43.16 & 47.68 & 17.48 & 34.67 & 12.42 & 43.60 & 34.68 & 36.67 & 43.43 \\ 
    &  & SEA & 38.11 & 38.52 & 30.00 & 42.26 & 51.88 & 38.91 & 0.00 & 40.61 & 26.77 & 45.46 & 30.01 & 35.02 & 35.62 \\ 
    & vi & ENG & 45.93 & 41.01 & 36.06 & 28.55 & 40.52 & 47.87 & 7.75 & 39.14 & 12.79 & 40.75 & 33.60 & 35.22 & 45.65 \\ 
    &  & SEA & 49.02 & 35.37 & 7.56 & 31.51 & 49.28 & 29.41 & 2.29 & 32.44 & 19.93 & 37.84 & 27.80 & 29.10 & 44.44 \\ 
    & zh & ENG & 26.98 & 27.05 & 30.02 & 21.80 & 24.01 & 28.18 & 13.23 & 26.33 & 11.63 & 27.03 & 21.28 & 36.11 & 42.77 \\ 
    &  & SEA & 29.36 & 24.67 & 26.57 & 29.41 & 30.63 & 27.65 & 34.61 & 26.28 & 28.19 & 22.09 & 18.45 & 33.22 & 33.09 \\ 
    \midrule
    \bf Average &  & ENG & 32.99 & 33.14 & 29.77 & 26.28 & 35.42 & 37.26 & 17.23 & 29.78 & 15.23 & 32.41 & 26.82 & 36.93 & 36.91 \\ 
    &  & SEA & 32.46 & 28.16 & 18.08 & 28.42 & 34.09 & 28.49 & 15.39 & 24.75 & 18.28 & 29.97 & 22.70 & 36.33 & 34.27 \\ 
    \bottomrule
    \end{tabular}}
\end{table*}

\begin{table*}[!ht]
    \centering
    \caption{Emotion Recognition performance: results across datasets comparing English and SEA prompts. Evaluation is conducted using the Model-as-Judge metric, where higher scores indicate better performance.}
    \label{tab:appendix_emotion}
    \resizebox{0.99\linewidth}{!}{%
    \renewcommand{\arraystretch}{1.5}
    \begin{tabular}{
        c@{\;}c@{\;}c@{\;}|
        *{12}{c@{\hspace{0.3em}}}
        c
        }
    \toprule
    \multicolumn{3}{c|}{\textbf{Model}}
    & \makecell{\textbf{MERaLiON2} \\ \textbf{10B}} 
    & \makecell{\textbf{MERaLiON2} \\ \textbf{3B}} 
    & \makecell{\textbf{SeaLLMs} \\ \textbf{Audio} \\ \textbf{7B}}
    & \makecell{\textbf{Phi-4} \\ \textbf{multi-} \\ \textbf{-modal} \\ \textbf{instruct}}
    & \makecell{\textbf{Kimi} \\ \textbf{Audio}} 
    & \makecell{\textbf{Voxtral} \\ \textbf{mini}} 
    & \makecell{\textbf{Qwen2} \\ \textbf{Audio} \\ \textbf{7B} \\ \textbf{Instruct}} 
    & \makecell{\textbf{Qwen} \\ \textbf{2.5} \\ \textbf{Omni} \\ \textbf{3B}}
    & \makecell{\textbf{Qwen} \\ \textbf{2.5} \\ \textbf{Omni} \\ \textbf{7B}}
    & \makecell{\textbf{gemma} \\ \textbf{3n} \\ \textbf{E4B-it}}
    & \makecell{\textbf{gemma} \\ \textbf{3n} \\ \textbf{E2B-it}}
    & \makecell{\textbf{Gemini} \\ \textbf{2.5} \\ \textbf{Flash}} 
    & \makecell{\textbf{GPT} \\ \textbf{4o} \\ \textbf{Audio}} \\ 
    \multicolumn{3}{c|}{Size} & 10B & 3B & 7B & 5.6B & 7B & 3B & 7B & 3B & 7B & 4B & 2B & - & - \\ 
    \midrule
    \textbf{Data} 
    & \textbf{Lang} 
    & \textbf{Prompt} 
    & & & & & & & & & & & & & \\ 
    \cmidrule{1-3}
    \bf EmoTa & ta & ENG & 8.33 & 15.12 & 6.78 & 21.58 & 15.33 & 8.33 & 23.08 & 7.26 & 14.58 & 8.49 & 8.23 & 12.00 & 8.00 \\
    &  & SEA & 11.97 & 10.84 & 0.85 & 2.24 & 7.91 & 3.73 & 0.96 & 0.85 & 1.00 & 6.94 & 11.43 & 9.00 & 17.00 \\
    \bf ESD & en & ENG & 19.50 & 23.40 & 12.85 & 19.80 & 64.70 & 9.05 & 40.90 & 9.40 & 13.55 & 8.50 & 9.15 & 15.00 & 13.00 \\
    &  & SEA & 19.50 & 23.40 & 12.85 & 19.80 & 64.70 & 9.05 & 40.90 & 9.40 & 13.55 & 8.50 & 9.15 & 15.00 & 13.00 \\
    & zh & ENG & 16.70 & 19.90 & 7.70 & 17.80 & 71.25 & 5.55 & 34.60 & 15.50 & 18.10 & 10.55 & 11.60 & 14.00 & 7.00 \\
    &  & SEA & 14.15 & 18.75 & 7.45 & 12.50 & 66.15 & 4.80 & 47.55 & 17.70 & 16.10 & 12.65 & 13.65 & 14.00 & 13.50 \\
    \bf IndoWaveSentiment & id & ENG & 20.00 & 20.67 & 11.67 & 18.33 & 27.83 & 11.00 & 13.33 & 11.33 & 13.33 & 17.00 & 15.83 & 26.00 & 23.00 \\
    &  & SEA & 23.33 & 18.00 & 17.50 & 3.67 & 29.83 & 8.67 & 15.50 & 16.00 & 14.33 & 13.33 & 9.67 & 18.00 & 22.00 \\
    \bf M3ED & zh & ENG & 19.35 & 26.65 & 10.90 & 20.65 & 19.00 & 7.95 & 14.30 & 11.95 & 13.50 & 7.35 & 10.20 & 16.00 & 12.00 \\
    &  & SEA & 23.25 & 23.40 & 13.35 & 13.50 & 23.15 & 0.85 & 19.80 & 18.55 & 19.60 & 13.20 & 10.75 & 16.50 & 14.00 \\
    \bf TEC & ta & ENG & 34.85 & 42.42 & 24.85 & 30.30 & 42.42 & 23.12 & 30.91 & 28.48 & 27.58 & 21.52 & 19.39 & 42.50 & 43.00 \\
    &  & SEA & 36.97 & 20.30 & 1.82 & 3.64 & 21.81 & 6.06 & 2.42 & 1.21 & 1.21 & 31.52 & 27.88 & 35.00 & 47.00 \\
    \bf THAI SER & th & ENG & 12.36 & 19.74 & 11.62 & 17.59 & 17.48 & 9.31 & 14.19 & 10.26 & 13.66 & 13.82 & 11.10 & 11.00 & 13.00 \\ 
    &  & SEA & 13.19 & 16.44 & 10.37 & 9.06 & 84.29 & 4.29 & 8.38 & 5.97 & 7.33 & 11.36 & 9.63 & 10.00 & 10.00 \\ \midrule
    \bf Average &  & ENG & 22.56 & 26.55 & 15.53 & 22.99 & 35.88 & 14.55 & 26.74 & 15.79 & 19.10 & 17.61 & 16.75 & 23.81 & 19.88 \\
    &  & SEA & 23.97 & 21.96 & 12.76 & 12.78 & 40.86 & 9.94 & 22.26 & 12.72 & 13.95 & 18.90 & 17.59 & 21.44 & 22.06 \\
    \bottomrule
    \end{tabular}}
\end{table*}

\clearpage

\begin{table*}[!ht]
    \centering
    \caption{Gender Recognition performance: results across datasets comparing English and SEA prompts. Evaluation is conducted using the macro-F1 metric ($\%$), where higher scores indicate better performance.}
    \label{tab:appendix_gender}
    \resizebox{0.99\linewidth}{!}{%
    \renewcommand{\arraystretch}{1.5}
    \begin{tabular}{
        c@{\;}c@{\;}c@{\;}|
        *{14}{c@{\hspace{0.3em}}}
        c
        }
    \toprule
    \multicolumn{3}{c|}{\textbf{Model}}
    & \makecell{\textbf{MERaLiON2} \\ \textbf{10B}} 
    & \makecell{\textbf{MERaLiON2} \\ \textbf{3B}} 
    & \makecell{\textbf{SeaLLMs} \\ \textbf{Audio} \\ \textbf{7B}}
    & \makecell{\textbf{Phi-4} \\ \textbf{multi-} \\ \textbf{-modal} \\ \textbf{instruct}}
    & \makecell{\textbf{Kimi} \\ \textbf{Audio}} 
    & \makecell{\textbf{Voxtral} \\ \textbf{mini}} 
    & \makecell{\textbf{Qwen2} \\ \textbf{Audio} \\ \textbf{7B} \\ \textbf{Instruct}} 
    & \makecell{\textbf{Qwen} \\ \textbf{2.5} \\ \textbf{Omni} \\ \textbf{3B}}
    & \makecell{\textbf{Qwen} \\ \textbf{2.5} \\ \textbf{Omni} \\ \textbf{7B}}
    & \makecell{\textbf{gemma} \\ \textbf{3n} \\ \textbf{E4B-it}}
    & \makecell{\textbf{gemma} \\ \textbf{3n} \\ \textbf{E2B-it}}
    & \makecell{\textbf{Gemini} \\ \textbf{2.5} \\ \textbf{Flash}} 
    & \makecell{\textbf{GPT} \\ \textbf{4o} \\ \textbf{Audio}} \\ 
    \multicolumn{3}{c|}{Size} & 10B & 3B & 7B & 5.6B & 7B & 3B & 7B & 3B & 7B & 4B & 2B & - & - \\ 
    \midrule
    \textbf{Data} 
    & \textbf{Lang} 
    & \textbf{Prompt} 
    & & & & & & & & & & & & & \\ 
    \cmidrule{1-3}
    \bf Commonvoice & id & ENG & 38.23 & 35.03 & 36.04 & 42.81 & 94.57 & 45.34 & 95.18 & 58.43 & 74.16 & 35.39 & 16.81 & 96.22 & 60.31 \\
    &  & SEA & 47.58 & 33.12 & 34.10 & 31.50 & 91.93 & 29.64 & 75.60 & 43.60 & 58.73 & 46.44 & 33.73 & 90.80 & 44.22 \\
    & ta & ENG & 52.73 & 38.37 & 33.56 & 48.13 & 92.88 & 44.83 & 96.80 & 38.12 & 70.49 & 26.61 & 10.54 & 94.47 & 46.74 \\
    &  & SEA & 41.64 & 34.85 & 4.79 & 10.66 & 33.61 & 21.47 & 25.32 & 12.92 & 19.96 & 11.10 & 2.48 & 85.25 & 41.71 \\
    & th & ENG & 50.07 & 29.87 & 38.95 & 35.31 & 96.93 & 54.20 & 96.76 & 62.84 & 78.78 & 32.59 & 12.22 & 93.48 & 64.01 \\
    &  & SEA & 20.81 & 4.76 & 44.18 & 11.63 & 67.71 & 2.96 & 82.51 & 61.01 & 71.88 & 13.94 & 3.87 & 87.45 & 53.81 \\
    & vi & ENG & 23.74 & 17.43 & 46.84 & 39.02 & 95.90 & 54.45 & 92.35 & 61.88 & 68.36 & 20.90 & 10.40 & 87.31 & 45.73 \\
    &  & SEA & 20.01 & 41.91 & 44.96 & 22.60 & 93.77 & 25.94 & 95.80 & 38.37 & 30.73 & 1.45 & 0.00 & 74.62 & 56.48 \\
    & zh & ENG & 52.79 & 30.31 & 40.79 & 50.41 & 90.54 & 48.97 & 97.93 & 83.60 & 86.13 & 34.45 & 6.85 & 88.43 & 62.54 \\
    &  & SEA & 37.20 & 34.72 & 40.79 & 43.52 & 74.76 & 34.59 & 98.16 & 75.63 & 89.31 & 22.57 & 0.64 & 86.88 & 26.77 \\
    \bf EmoTa & ta & ENG & 65.54 & 36.64 & 34.80 & 50.12 & 93.67 & 39.09 & 98.82 & 30.27 & 55.79 & 25.64 & 11.17 & 95.44 & 39.70 \\
    &  & SEA & 47.91 & 38.67 & 3.13 & 11.39 & 47.22 & 18.68 & 13.59 & 11.69 & 21.08 & 17.71 & 4.16 & 83.62 & 41.39 \\
    \bf FLEURS & en & ENG & 59.61 & 66.86 & 31.89 & 65.92 & 98.55 & 19.53 & 99.33 & 55.32 & 41.41 & 46.93 & 10.24 & 97.51 & 85.37 \\
    &  & SEA & 59.61 & 66.86 & 31.89 & 65.92 & 98.55 & 19.53 & 99.33 & 55.32 & 41.41 & 46.93 & 10.24 & 97.51 & 85.37 \\
    & km & ENG & 53.66 & 30.84 & 44.04 & 31.43 & 99.72 & 52.10 & 77.08 & 52.26 & 63.85 & 52.93 & 3.90 & 98.91 & 76.54 \\
    &  & SEA & 22.28 & 42.15 & 3.02 & 20.22 & 24.52 & 22.09 & 16.94 & 19.85 & 11.66 & 1.77 & 0.86 & 39.02 & 24.79 \\
    \bf IndoWaveSentiment & id & ENG & 71.01 & 63.16 & 34.07 & 51.59 & 91.74 & 15.75 & 98.33 & 22.75 & 78.39 & 34.81 & 11.04 & 96.82 & 73.65 \\
    &  & SEA & 64.48 & 32.56 & 33.33 & 39.22 & 81.73 & 32.78 & 93.98 & 41.83 & 66.68 & 41.24 & 30.43 & 93.68 & 50.52 \\
    \bf M3ED & zh & ENG & 84.97 & 59.73 & 36.53 & 84.32 & 82.25 & 16.68 & 86.26 & 26.40 & 59.61 & 33.17 & 9.50 & 85.32 & 36.71 \\
    &  & SEA & 74.30 & 80.58 & 42.16 & 64.81 & 83.31 & 11.55 & 77.36 & 20.49 & 82.63 & 14.87 & 0.20 & 68.99 & 17.73 \\
    \bf OpenSLR & ta & ENG & 52.05 & 44.09 & 34.00 & 47.59 & 97.84 & 40.65 & 99.10 & 58.42 & 58.66 & 23.74 & 8.79 & 96.50 & 57.92 \\
    &  & SEA & 42.57 & 32.81 & 5.30 & 8.96 & 46.66 & 17.99 & 27.84 & 14.33 & 20.92 & 12.67 & 4.06 & 89.78 & 41.43 \\
    \bf SG Streets & en & ENG & 90.14 & 30.77 & 38.27 & 58.50 & 98.13 & 20.31 & 84.40 & 18.11 & 18.38 & 27.36 & 17.87 & 97.80 & 49.63 \\
    &  & SEA & 90.14 & 30.77 & 38.27 & 58.50 & 98.13 & 20.31 & 84.40 & 18.11 & 18.38 & 27.36 & 17.87 & 97.80 & 49.63 \\
    \bf ASR-SMalDuSC & ms & ENG & 52.00 & 34.45 & 38.20 & 31.49 & 96.84 & 46.97 & 93.89 & 78.48 & 72.21 & 41.14 & 13.62 & 98.00 & 83.01 \\
    &  & SEA & 50.16 & 38.35 & 38.58 & 47.51 & 94.72 & 35.38 & 54.57 & 51.03 & 53.16 & 59.03 & 36.58 & 95.88 & 57.88 \\
    \bf Thai Elderly Speech & th & ENG & 49.98 & 38.30 & 29.48 & 30.44 & 94.69 & 58.52 & 97.01 & 62.45 & 75.18 & 39.56 & 7.83 & 98.86 & 56.69 \\
    &  & SEA & 19.20 & 7.83 & 44.34 & 9.52 & 65.42 & 4.27 & 78.53 & 54.76 & 81.96 & 21.08 & 6.11 & 82.87 & 63.12 \\
    \bf THAI SER & th & ENG & 58.52 & 60.45 & 39.30 & 65.04 & 87.70 & 41.09 & 87.41 & 44.51 & 74.55 & 42.46 & 13.75 & 86.95 & 58.33 \\
    &  & SEA & 33.53 & 0.99 & 81.54 & 11.89 & 69.81 & 14.25 & 66.51 & 48.69 & 81.08 & 24.90 & 7.05 & 75.95 & 47.47 \\
    \bf Vietnam-Celeb & vi & ENG & 64.95 & 46.66 & 36.44 & 57.59 & 71.65 & 38.03 & 69.56 & 43.74 & 64.45 & 48.93 & 19.35 & 68.64 & 53.59 \\
    &  & SEA & 63.43 & 62.72 & 34.00 & 39.13 & 69.87 & 15.17 & 69.33 & 29.15 & 37.18 & 13.51 & 2.53 & 70.45 & 46.22 \\ \midrule
    \bf Average &  & ENG & 57.50 & 41.44 & 37.07 & 49.36 & 92.73 & 39.78 & 91.89 & 49.85 & 65.03 & 35.41 & 11.49 & 92.54 & 59.40 \\
    &  & SEA & 45.93 & 36.48 & 32.77 & 31.06 & 71.36 & 20.41 & 66.24 & 37.30 & 49.17 & 23.54 & 10.05 & 82.54 & 46.78 \\
    \bottomrule
    \end{tabular}}
\end{table*}

\clearpage

\begin{table*}[!ht]
    \centering
    \caption{Speaker Recognition performance: results across datasets comparing English and SEA prompts. Evaluation is conducted using the macro-F1 metric ($\%$), where higher scores indicate better performance.}
    \label{tab:appendix_speaker}
    \resizebox{0.99\linewidth}{!}{%
    \renewcommand{\arraystretch}{1.5}
    \begin{tabular}{
        c@{\;}c@{\;}c@{\;}|
        *{12}{c@{\hspace{0.3em}}}
        c
        }
    \toprule
    \multicolumn{3}{c|}{\textbf{Model}}
    & \makecell{\textbf{MERaLiON2} \\ \textbf{10B}} 
    & \makecell{\textbf{MERaLiON2} \\ \textbf{3B}} 
    & \makecell{\textbf{SeaLLMs} \\ \textbf{Audio} \\ \textbf{7B}}
    & \makecell{\textbf{Phi-4} \\ \textbf{multi-} \\ \textbf{-modal} \\ \textbf{instruct}}
    & \makecell{\textbf{Kimi} \\ \textbf{Audio}} 
    & \makecell{\textbf{Voxtral} \\ \textbf{mini}} 
    & \makecell{\textbf{Qwen2} \\ \textbf{Audio} \\ \textbf{7B} \\ \textbf{Instruct}} 
    & \makecell{\textbf{Qwen} \\ \textbf{2.5} \\ \textbf{Omni} \\ \textbf{3B}}
    & \makecell{\textbf{Qwen} \\ \textbf{2.5} \\ \textbf{Omni} \\ \textbf{7B}}
    & \makecell{\textbf{gemma} \\ \textbf{3n} \\ \textbf{E4B-it}}
    & \makecell{\textbf{gemma} \\ \textbf{3n} \\ \textbf{E2B-it}}
    & \makecell{\textbf{Gemini} \\ \textbf{2.5} \\ \textbf{Flash}} 
    & \makecell{\textbf{GPT} \\ \textbf{4o} \\ \textbf{Audio}} \\ 
    \multicolumn{3}{c|}{Size} & 10B & 3B & 7B & 5.6B & 7B & 3B & 7B & 3B & 7B & 4B & 2B & - & - \\ 
    \midrule
    \textbf{Data} 
    & \textbf{Lang} 
    & \textbf{Prompt} 
    & & & & & & & & & & & & & \\ 
    \cmidrule{1-3}
    \bf EmoTa & ta & ENG & 50.08 & 45.03 & 45.79 & 27.16 & 49.96 & 42.09 & 46.52 & 22.49 & 14.57 & 34.98 & 34.79 & 59.35 & 10.59 \\
    &  & SEA & 34.44 & 32.75 & 23.51 & 24.85 & 17.77 & 29.37 & 26.78 & 25.34 & 23.62 & 26.60 & 35.77 & 41.23 & 23.10 \\
    \bf ESD & en & ENG & 46.35 & 43.57 & 48.21 & 48.48 & 48.15 & 42.01 & 41.28 & 21.77 & 7.55 & 31.12 & 36.79 & 60.54 & 14.96 \\
    &  & SEA & 46.35 & 43.57 & 48.21 & 48.48 & 48.15 & 42.01 & 41.28 & 21.77 & 7.55 & 31.12 & 36.79 & 60.54 & 14.96 \\
    & zh & ENG & 48.13 & 48.70 & 42.93 & 31.57 & 42.43 & 40.26 & 41.28 & 43.06 & 26.15 & 34.25 & 36.69 & 48.16 & 15.74 \\
    &  & SEA & 51.58 & 44.06 & 35.55 & 32.33 & 39.67 & 38.79 & 40.44 & 38.55 & 40.08 & 39.91 & 35.14 & 45.15 & 32.66 \\
    \bf mig & my & ENG & 47.31 & 42.60 & 38.93 & 39.11 & 48.44 & 39.87 & 46.35 & 23.83 & 15.71 & 30.26 & 31.28 & 54.61 & 1.96 \\
    &  & SEA & 22.65 & 8.31 & 4.70 & 31.71 & 31.39 & 33.11 & 6.98 & 25.93 & 18.88 & 32.83 & 30.95 & 39.29 & 25.11 \\
    \bf ASR-SMalDuSC & ms & ENG & 53.35 & 42.97 & 47.34 & 35.41 & 45.35 & 43.17 & 38.46 & 19.58 & 13.94 & 31.92 & 37.05 & 57.77 & 21.56 \\
    &  & SEA & 42.16 & 33.83 & 32.01 & 22.33 & 24.72 & 47.21 & 17.87 & 35.61 & 24.31 & 34.28 & 28.41 & 71.60 & 18.95 \\
    \bf Thai Elderly Speech & th & ENG & 48.31 & 43.56 & 45.73 & 36.62 & 44.01 & 44.21 & 41.51 & 22.45 & 12.60 & 34.00 & 38.37 & 66.85 & 10.88 \\
    &  & SEA & 28.01 & 24.64 & 32.24 & 35.31 & 17.29 & 47.32 & 33.09 & 38.04 & 11.46 & 34.12 & 23.71 & 55.48 & 17.74 \\
    \bf THAI SER & th & ENG & 45.87 & 41.34 & 38.30 & 35.07 & -- & 48.22 & 39.37 & 23.52 & 18.91 & 33.76 & 41.13 & 53.90 & 27.99 \\
    &  & SEA & 28.29 & 26.65 & 31.49 & 37.88 & 24.33 & 39.05 & 39.72 & 30.62 & 9.64 & 35.90 & 27.36 & 37.86 & 10.59 \\
    \bf VoxVietnam-O & vi & ENG & 47.52 & 39.86 & 40.76 & 37.98 & 47.73 & 42.86 & 40.65 & 25.08 & 14.95 & 33.82 & 41.20 & 62.93 & 12.23 \\
    &  & SEA & 43.11 & 25.44 & 40.67 & 28.96 & 33.58 & 32.54 & 47.09 & 35.89 & 13.33 & 38.28 & 38.45 & 61.19 & 9.17 \\ \midrule
    \bf Average &  & ENG & 48.36 & 43.45 & 43.50 & 36.43 & 46.58 & 42.84 & 41.93 & 25.22 & 15.55 & 33.01 & 37.16 & 58.01 & 14.49 \\
    &  & SEA & 37.07 & 29.91 & 31.05 & 32.73 & 29.61 & 38.67 & 31.66 & 31.47 & 18.61 & 34.13 & 32.07 & 51.54 & 19.04 \\
    \bottomrule
    \end{tabular}}
\end{table*}

\begin{table*}[!ht]
    \centering
    \caption{Speech Translation performance: results across datasets comparing English and SEA prompts. Evaluation is based on the BLEU metric, where higher scores indicate better performance.}
    \label{tab:appendix_st_bleu}
    \resizebox{0.99\linewidth}{!}{%
    \renewcommand{\arraystretch}{1.5}
    \begin{tabular}{
        c@{\;}c@{\;}c@{\;}|
        *{12}{c@{\hspace{0.3em}}}
        c
        }
    \toprule
    \multicolumn{3}{c|}{\textbf{Model}}
    & \makecell{\textbf{MERaLiON2} \\ \textbf{10B}} 
    & \makecell{\textbf{MERaLiON2} \\ \textbf{3B}} 
    & \makecell{\textbf{SeaLLMs} \\ \textbf{Audio} \\ \textbf{7B}}
    & \makecell{\textbf{Phi-4} \\ \textbf{multi-} \\ \textbf{-modal} \\ \textbf{instruct}}
    & \makecell{\textbf{Kimi} \\ \textbf{Audio}} 
    & \makecell{\textbf{Voxtral} \\ \textbf{mini}} 
    & \makecell{\textbf{Qwen2} \\ \textbf{Audio} \\ \textbf{7B} \\ \textbf{Instruct}} 
    & \makecell{\textbf{Qwen} \\ \textbf{2.5} \\ \textbf{Omni} \\ \textbf{3B}}
    & \makecell{\textbf{Qwen} \\ \textbf{2.5} \\ \textbf{Omni} \\ \textbf{7B}}
    & \makecell{\textbf{gemma} \\ \textbf{3n} \\ \textbf{E4B-it}}
    & \makecell{\textbf{gemma} \\ \textbf{3n} \\ \textbf{E2B-it}}
    & \makecell{\textbf{Gemini} \\ \textbf{2.5} \\ \textbf{Flash}} 
    & \makecell{\textbf{GPT} \\ \textbf{4o} \\ \textbf{Audio}} \\ 
    \multicolumn{3}{c|}{Size} & 10B & 3B & 7B & 5.6B & 7B & 3B & 7B & 3B & 7B & 4B & 2B & - & - \\ 
    \midrule
    \textbf{Data} 
    & \textbf{Lang} 
    & \textbf{Prompt} 
    & & & & & & & & & & & & & \\ 
    \cmidrule{1-3}
    \bf FLEURS & id & ENG & 32.48 & 20.97 & 25.55 & 0.70 & 10.75 & 35.02 & 9.37 & 16.65 & 16.55 & 21.40 & 20.52 & 24.07 & 33.44 \\
    &  & SEA & 30.02 & 21.98 & 25.97 & 0.13 & 16.81 & 34.58 & 6.23 & 11.15 & 17.93 & 24.80 & 17.57 & 24.43 & 34.90 \\
    & km & ENG & 2.19 & 0.50 & 0.73 & 0.23 & 0.35 & 7.04 & 0.47 & 0.26 & 0.48 & 4.72 & 2.52 & 12.40 & 6.34 \\
    &  & SEA & 1.97 & 0.45 & 0.33 & 0.05 & 0.49 & 0.08 & 0.11 & 0.27 & 0.24 & 4.71 & 2.50 & 13.76 & 7.52 \\
    & lo & ENG & 11.34 & 1.03 & 5.91 & 0.04 & 0.55 & 10.44 & 0.47 & 2.49 & 3.06 & 7.66 & 6.75 & 15.36 & 13.61 \\
    &  & SEA & 12.09 & 0.88 & 1.04 & 0.01 & 1.04 & 0.55 & 0.11 & 0.80 & 2.35 & 8.94 & 7.18 & 15.04 & 13.06 \\
    & ms & ENG & 31.04 & 15.52 & 19.71 & 1.04 & 6.15 & 30.44 & 5.25 & 12.16 & 13.19 & 18.24 & 16.41 & 26.49 & 33.53 \\
    &  & SEA & 34.35 & 13.09 & 20.30 & 0.22 & 20.02 & 30.33 & 2.57 & 10.12 & 14.24 & 23.04 & 14.37 & 23.70 & 33.93 \\
    & my & ENG & 0.37 & 0.12 & 0.29 & 0.11 & 0.03 & 0.73 & 0.40 & 0.11 & 0.35 & 0.78 & 0.35 & 8.59 & 1.47 \\
    &  & SEA & 0.59 & 0.14 & 0.03 & 0.11 & 0.04 & 0.63 & 0.14 & 0.02 & 0.10 & 0.73 & 0.10 & 15.05 & 2.08 \\
    & th & ENG & 17.22 & 3.47 & 12.30 & 0.14 & 2.42 & 19.53 & 0.47 & 7.90 & 8.34 & 12.95 & 9.57 & 16.93 & 22.41 \\
    &  & SEA & 18.70 & 5.59 & 11.51 & 0.02 & 3.52 & 20.15 & 0.55 & 7.86 & 9.29 & 14.90 & 8.49 & 20.25 & 22.93 \\
    & tl & ENG & 25.79 & 8.28 & 1.67 & 0.92 & 5.99 & 30.42 & 1.68 & 2.52 & 2.50 & 17.14 & 12.98 & 14.50 & 28.48 \\
    &  & SEA & 27.22 & 7.76 & 1.76 & 0.10 & 23.27 & 30.69 & 1.25 & 0.89 & 2.29 & 22.47 & 14.44 & 16.09 & 26.20 \\
    & vi & ENG & 19.18 & 6.50 & 13.52 & 0.12 & 3.94 & 24.79 & 2.05 & 11.40 & 12.02 & 7.19 & 4.89 & 17.26 & 27.10 \\
    &  & SEA & 22.91 & 5.95 & 13.00 & 0.01 & 4.19 & 25.05 & 1.91 & 10.31 & 12.32 & 11.84 & 7.51 & 19.74 & 30.27 \\
    & zh & ENG & 20.14 & 11.63 & 17.01 & 24.08 & 0.01 & 21.36 & 20.69 & 14.89 & 14.73 & 8.72 & 6.75 & 16.18 & 24.77 \\
    &  & SEA & 20.74 & 10.94 & 17.11 & 2.28 & 0.01 & 21.30 & 15.88 & 15.02 & 13.81 & 10.78 & 6.37 & 21.91 & 21.60 \\ \midrule
    \bf Average &  & ENG & 17.75 & 7.56 & 10.74 & 3.04 & 3.36 & 19.98 & 4.54 & 7.60 & 7.91 & 10.98 & 8.97 & 16.86 & 21.24 \\
    &  & SEA & 18.73 & 7.42 & 10.11 & 0.32 & 7.71 & 18.15 & 3.20 & 6.27 & 8.06 & 13.58 & 8.72 & 18.89 & 21.39 \\
    \bottomrule
    \end{tabular}}
\end{table*}

\begin{table*}[!ht]
    \centering
    \caption{Speech Translation performance: results across datasets comparing English and SEA prompts. Evaluation is based on the chrF metric, where higher scores indicate better performance.}
    \label{tab:appendix_st_chrf}
    \resizebox{0.99\linewidth}{!}{%
    \renewcommand{\arraystretch}{1.5}
    \begin{tabular}{
        c@{\;}c@{\;}c@{\;}|
        *{12}{c@{\hspace{0.3em}}}
        c
        }
    \toprule
    \multicolumn{3}{c|}{\textbf{Model}}
    & \makecell{\textbf{MERaLiON2} \\ \textbf{10B}} 
    & \makecell{\textbf{MERaLiON2} \\ \textbf{3B}} 
    & \makecell{\textbf{SeaLLMs} \\ \textbf{Audio} \\ \textbf{7B}}
    & \makecell{\textbf{Phi-4} \\ \textbf{multi-} \\ \textbf{-modal} \\ \textbf{instruct}}
    & \makecell{\textbf{Kimi} \\ \textbf{Audio}} 
    & \makecell{\textbf{Voxtral} \\ \textbf{mini}} 
    & \makecell{\textbf{Qwen2} \\ \textbf{Audio} \\ \textbf{7B} \\ \textbf{Instruct}} 
    & \makecell{\textbf{Qwen} \\ \textbf{2.5} \\ \textbf{Omni} \\ \textbf{3B}}
    & \makecell{\textbf{Qwen} \\ \textbf{2.5} \\ \textbf{Omni} \\ \textbf{7B}}
    & \makecell{\textbf{gemma} \\ \textbf{3n} \\ \textbf{E4B-it}}
    & \makecell{\textbf{gemma} \\ \textbf{3n} \\ \textbf{E2B-it}}
    & \makecell{\textbf{Gemini} \\ \textbf{2.5} \\ \textbf{Flash}} 
    & \makecell{\textbf{GPT} \\ \textbf{4o} \\ \textbf{Audio}} \\ 
    \multicolumn{3}{c|}{Size} & 10B & 3B & 7B & 5.6B & 7B & 3B & 7B & 3B & 7B & 4B & 2B & - & - \\ 
    \midrule
    \textbf{Data} 
    & \textbf{Lang} 
    & \textbf{Prompt} 
    & & & & & & & & & & & & & \\ 
    \cmidrule{1-3}
    \bf FLEURS & id & ENG & 57.23 & 47.58 & 50.76 & 15.81 & 33.74 & 59.29 & 34.12 & 50.49 & 53.81 & 43.48 & 44.25 & 52.72 & 61.14 \\
    &  & SEA & 62.44 & 49.18 & 51.20 & 8.98 & 44.94 & 59.13 & 27.48 & 36.68 & 55.30 & 55.49 & 46.84 & 59.12 & 62.00 \\
    & km & ENG & 24.72 & 18.37 & 15.85 & 10.73 & 8.87 & 29.23 & 15.36 & 16.19 & 17.56 & 27.44 & 24.97 & 42.54 & 32.21 \\
    &  & SEA & 25.22 & 19.05 & 11.14 & 4.80 & 9.41 & 0.13 & 14.69 & 17.82 & 16.95 & 28.97 & 23.53 & 49.25 & 31.58 \\
    & lo & ENG & 39.13 & 17.32 & 29.48 & 9.85 & 11.00 & 34.78 & 13.75 & 22.68 & 25.75 & 27.20 & 25.10 & 46.14 & 41.65 \\
    &  & SEA & 39.87 & 19.81 & 7.79 & 0.38 & 13.54 & 4.27 & 14.29 & 20.12 & 24.56 & 36.27 & 29.32 & 51.11 & 40.16 \\
    & ms & ENG & 56.32 & 40.29 & 45.47 & 17.25 & 28.25 & 55.49 & 28.11 & 43.16 & 47.32 & 41.49 & 41.66 & 54.32 & 59.00 \\
    &  & SEA & 60.15 & 37.20 & 45.91 & 9.18 & 44.62 & 55.51 & 21.38 & 38.21 & 46.72 & 51.09 & 40.56 & 57.81 & 59.13 \\
    & my & ENG & 16.85 & 9.27 & 14.41 & 9.99 & 3.00 & 15.65 & 15.70 & 15.39 & 16.91 & 18.92 & 15.82 & 34.38 & 22.55 \\
    &  & SEA & 20.16 & 12.58 & 4.30 & 3.64 & 3.26 & 15.08 & 9.68 & 5.32 & 15.15 & 16.94 & 4.83 & 48.55 & 23.20 \\
    & th & ENG & 45.30 & 18.67 & 38.09 & 10.64 & 23.01 & 46.16 & 13.66 & 36.28 & 40.44 & 37.96 & 33.85 & 46.55 & 51.64 \\
    &  & SEA & 48.45 & 28.54 & 38.14 & 1.03 & 27.79 & 46.35 & 14.20 & 34.22 & 41.07 & 42.99 & 29.54 & 54.86 & 52.94 \\
    & tl & ENG & 53.10 & 33.65 & 18.72 & 15.40 & 28.59 & 53.79 & 18.93 & 22.54 & 25.28 & 44.26 & 38.89 & 44.36 & 57.98 \\
    &  & SEA & 53.92 & 32.81 & 18.08 & 6.02 & 48.67 & 53.94 & 15.79 & 19.01 & 23.48 & 51.68 & 43.97 & 55.00 & 57.26 \\
    & vi & ENG & 44.70 & 29.51 & 37.19 & 11.46 & 25.51 & 49.57 & 23.04 & 42.28 & 45.73 & 26.37 & 24.41 & 42.79 & 54.26 \\
    &  & SEA & 49.76 & 27.54 & 37.40 & 4.12 & 26.72 & 49.66 & 21.03 & 37.99 & 45.60 & 37.58 & 30.90 & 54.37 & 56.11 \\
    & zh & ENG & 49.39 & 40.57 & 41.94 & 52.73 & 0.19 & 47.70 & 47.65 & 47.96 & 50.04 & 31.16 & 29.90 & 42.38 & 54.23 \\
    &  & SEA & 50.11 & 36.25 & 41.67 & 5.70 & 0.18 & 47.65 & 36.62 & 46.76 & 49.30 & 37.81 & 27.36 & 55.39 & 52.89 \\ \midrule
    \bf Average &  & ENG & 42.97 & 28.36 & 32.44 & 17.10 & 18.02 & 43.52 & 23.37 & 33.00 & 35.87 & 33.14 & 30.98 & 45.13 & 48.30 \\
    &  & SEA & 45.56 & 29.22 & 28.40 & 4.87 & 24.35 & 36.86 & 19.46 & 28.46 & 35.35 & 39.87 & 30.76 & 53.94 & 48.36 \\
    \bottomrule
    \end{tabular}}
\end{table*}

\clearpage
\begin{table*}[!ht]
    \centering
    \caption{TCQ performance: results across datasets comparing English and SEA prompts. Scores are reported as raw WER and CER values without  normalization; lower values indicate better performance.}
    \label{tab:appendix_tcq}
    \resizebox{0.88\linewidth}{!}{%
    \renewcommand{\arraystretch}{1.45}
    \begin{tabular}{
        c@{\;}c@{\;}c@{\;}c@{\;}|
        *{12}{c@{\hspace{0.3em}}}
        c
        }
    \toprule
    \multicolumn{4}{c|}{\textbf{Model}}
    & \makecell{\textbf{MERaLiON2} \\ \textbf{10B}} 
    & \makecell{\textbf{MERaLiON2} \\ \textbf{3B}} 
    & \makecell{\textbf{SeaLLMs} \\ \textbf{Audio} \\ \textbf{7B}}
    & \makecell{\textbf{Phi-4} \\ \textbf{multi-} \\ \textbf{-modal} \\ \textbf{instruct}}
    & \makecell{\textbf{Kimi} \\ \textbf{Audio}} 
    & \makecell{\textbf{Voxtral} \\ \textbf{mini}} 
    & \makecell{\textbf{Qwen2} \\ \textbf{Audio} \\ \textbf{7B} \\ \textbf{Instruct}} 
    & \makecell{\textbf{Qwen} \\ \textbf{2.5} \\ \textbf{Omni} \\ \textbf{3B}}
    & \makecell{\textbf{Qwen} \\ \textbf{2.5} \\ \textbf{Omni} \\ \textbf{7B}}
    & \makecell{\textbf{gemma} \\ \textbf{3n} \\ \textbf{E4B-it}}
    & \makecell{\textbf{gemma} \\ \textbf{3n} \\ \textbf{E2B-it}}
    & \makecell{\textbf{Gemini} \\ \textbf{2.5} \\ \textbf{Flash}} 
    & \makecell{\textbf{GPT} \\ \textbf{4o} \\ \textbf{Audio}} \\ 
    \multicolumn{4}{c|}{Size} & 10B & 3B & 7B & 5.6B & 7B & 3B & 7B & 3B & 7B & 4B & 2B & - & - \\ 
    \midrule
    \textbf{Data} 
    & \textbf{Lang}
    & \textbf{Length (Max)}
    & \textbf{Prompt} 
    & & & & & & & & & & & & & \\ 
    \cmidrule{1-4}
    \bf SG Streets & en & 30 & ENG & 2.43 & 2.43 & 2.12 & 2.67 & 2.47 & 1.89 & 2.30 & 2.64 & 4.21 & 2.20 & 2.32 & 0.47 & 2.24 \\
    & & & SEA & 2.43 & 2.43 & 2.12 & 2.67 & 2.47 & 1.89 & 2.30 & 2.64 & 4.21 & 2.20 & 2.32 & 0.47 & 2.24 \\
    &  & 60 & ENG & 3.44 & 3.47 & - & 2.10 & 3.36 & 2.12 & - & 2.29 & 3.35 & 1.91 & 1.94 & 0.76 & 2.49 \\
    & & & SEA & 3.44 & 3.47 & - & 2.10 & 3.36 & 2.12 & - & 2.29 & 3.35 & 1.91 & 1.94 & 0.76 & 2.49 \\
    \bf ASR-SgpCCSC & zh & 30 & ENG & 4.78 & 5.44 & 6.54 & 5.48 & 9.94 & 6.84 & 9.08 & 3.91 & 3.69 & 8.84 & 9.55 & 1.57 & 5.17 \\
    & & & SEA & 4.34 & 5.13 & 5.68 & 4.07 & 10.43 & 5.46 & 8.87 & 4.15 & 2.91 & 8.62 & 9.67 & 1.57 & 4.99 \\
    &  & 60 & ENG & 9.85 & 11.28 & - & 6.72 & 22.74 & 12.86 & - & 3.85 & 5.73 & 9.98 & 9.36 & 1.89 & 8.00 \\
    & & & SEA & 9.50 & 10.25 & - & 4.03 & 20.95 & 8.16 & - & 3.77 & 3.71 & 9.08 & 8.83 & 1.89 & 9.21 \\
    &  & 120 & ENG & 20.93 & 21.15 & - & 18.92 & 41.07 & 27.07 & - & - & 18.59 & - & - & 3.10 & 8.68 \\
    & & & SEA & 20.86 & 17.09 & - & 14.94 & 37.36 & 12.12 & - & - & 8.19 & - & - & 3.10 & 10.81 \\
    &  & 180 & ENG & 20.59 & - & - & - & 57.41 & 36.85 & - & - & - & - & - & 3.30 & 7.35 \\
    & & & SEA & 21.81 & - & - & - & 56.92 & 16.92 & - & - & - & - & - & 3.30 & 9.04 \\
    \bf YODAS2 & en & 30 & ENG & 5.34 & 5.31 & 3.81 & 4.53 & 5.79 & 3.62 & 4.57 & 4.94 & 4.42 & 5.03 & 5.03 & 1.36 & 4.42 \\
    & & & SEA & 5.34 & 5.31 & 3.81 & 4.53 & 5.79 & 3.62 & 4.57 & 4.94 & 4.42 & 5.03 & 5.03 & 1.36 & 4.42 \\
    &  & 60 & ENG & 10.98 & 10.74 & - & 6.39 & 11.76 & 5.48 & - & 7.01 & 4.86 & 5.83 & 5.96 & 1.29 & 6.37 \\
    & & & SEA & 10.98 & 10.74 & - & 6.39 & 11.76 & 5.48 & - & 7.01 & 4.86 & 5.83 & 5.96 & 1.29 & 6.37 \\
    &  & 120 & ENG & 16.19 & 15.10 & - & 9.02 & 22.28 & 9.09 & - & - & 6.92 & - & - & 1.82 & 6.49 \\
    & & & SEA & 16.19 & 15.10 & - & 9.02 & 22.28 & 9.09 & - & - & 6.92 & - & - & 1.82 & 6.49 \\
    &  & 180 & ENG & - & - & - & - & - & - & - & - & - & - & - & 1.75 & 6.72 \\
    & & & SEA & - & - & - & - & - & - & - & - & - & - & - & 1.75 & 6.72 \\
    & id & 30 & ENG & 3.82 & 3.70 & 3.42 & 21.41 & 22.32 & 2.71 & 4.05 & 3.76 & 3.23 & 3.78 & 3.81 & 0.64 & 3.18 \\
    & & & SEA & 3.77 & 3.50 & 3.52 & 14.13 & 9.58 & 3.06 & 3.86 & 3.81 & 4.54 & 4.00 & 3.91 & 1.36 & 3.20 \\
    &  & 60 & ENG & 7.88 & 6.44 & - & 14.18 & 19.97 & 4.40 & - & 5.26 & 4.09 & 4.23 & 4.30 & 1.21 & 4.47 \\
    & & & SEA & 7.63 & 6.00 & - & 9.43 & 8.76 & 4.75 & - & 5.53 & 6.53 & 4.24 & 4.44 & 1.79 & 5.23 \\
    &  & 120 & ENG & 13.07 & 9.45 & - & 13.92 & 16.49 & 7.27 & - & - & 6.09 & - & - & 1.38 & 4.88 \\
    & & & SEA & 12.59 & 9.08 & - & 12.35 & 10.94 & 6.27 & - & & 8.20 & - & - & 2.37 & 4.89 \\
    &  & 180 & ENG & 14.00 & - & - & - & 16.24 & 10.26 & - & - & - & - & - & 2.02 & 5.42 \\
    & & & SEA & 13.86 & - & - & - & 9.26 & 7.29 & - & - & - & - & - & 2.41 & 4.94 \\
    & th & 30 & ENG & 3.82 & 1.49 & 5.31 & 80.08 & 27.81 & 5.24 & 3.14 & 36.29 & 9.68 & 9.21 & 10.49 & 10.15 & 7.09 \\
    & & & SEA & 3.75 & 1.57 & 6.33 & 40.13 & 16.43 & 6.93 & 2.47 & 17.47 & 12.56 & 9.85 & 13.22 & 8.17 & 8.11 \\
    &  & 60 & ENG & 6.05 & 1.77 & - & 51.66 & 37.37 & 8.21 & - & 36.06 & 11.39 & 9.79 & 9.80 & 9.32 & 10.31 \\
    & & & SEA & 5.63 & 1.73 & - & 37.39 & 17.88 & 9.97 & - & 23.94 & 13.33 & 9.74 & 9.89 & 7.43 & 11.61 \\
    &  & 120 & ENG & 10.10 & 2.10 & - & 41.25 & 32.92 & 12.71 & - & - & 11.80 & - & - & 10.16 & 11.10 \\
    & & & SEA & 8.87 & 1.86 & - & 37.47 & 19.98 & 14.27 & - & & 14.77 & - & - & 8.02 & 11.84 \\
    &  & 180 & ENG & 10.52 & - & - & - & 60.78 & 14.44 & - & - & - & - & - & 12.21 & 12.98 \\
    & & & SEA & 9.76 & - & - & - & 27.66 & 16.71 & - & - & - & - & - & 9.93 & 12.87 \\
    & vi & 30 & ENG & 5.48 & 5.22 & 4.11 & 21.25 & 17.42 & 4.28 & 4.42 & 3.06 & 3.11 & 5.30 & 5.95 & 1.26 & 4.33 \\
    & & & SEA & 5.46 & 4.92 & 4.32 & 14.75 & 19.24 & 4.72 & 4.68 & 3.50 & 5.37 & 5.46 & 6.21 & 1.39 & 4.29 \\
    &  & 60 & ENG & 10.60 & 7.46 & - & 12.86 & 15.43 & 7.54 & - & 3.80 & 3.78 & 5.79 & 5.85 & 1.12 & 5.16 \\
    & & & SEA & 10.42 & 6.25 & - & 11.12 & 17.28 & 7.88 & - & 3.73 & 6.13 & 5.89 & 5.84 & 1.46 & 5.90 \\
    &  & 120 & ENG & 14.05 & 8.90 & - & 15.76 & 11.24 & 11.77 & - & - & 4.81 & - & - & 46.66 & 5.90 \\
    & & & SEA & 13.95 & 7.34 & - & 19.65 & 23.72 & 11.90 & - & - & 6.45 & - & - & 2.01 & 5.72 \\
    &  & 180 & ENG & 12.22 & - & - & - & 17.52 & 16.12 & - & - & - & - & - & 1.77 & 5.31 \\
    & & & SEA & 12.03 & - & - & - & 26.28 & 12.47 & - & - & - & - & - & 2.06 & 4.55 \\
    & zh & 30 & ENG & 10.16 & 9.83 & 13.04 & 9.29 & 12.21 & 8.61 & 11.38 & 6.57 & 10.10 & 12.95 & 14.12 & 1.41 & 8.99 \\
    & & & SEA & 8.46 & 9.40 & 11.19 & 8.49 & 12.23 & 6.91 & 11.00 & 5.31 & 7.93 & 12.94 & 13.58 & 2.98 & 10.74 \\
    &  & 60 & ENG & 17.60 & 16.31 & - & 10.41 & 22.87 & 13.54 & - & 6.15 & 12.89 & 14.16 & 15.26 & 2.86 & 10.92 \\
    & & & SEA & 15.95 & 14.34 & - & 11.48 & 23.07 & 8.33 & - & 4.67 & 9.09 & 13.88 & 15.08 & 4.40 & 13.01 \\
    &  & 120 & ENG & 28.66 & 22.18 & - & 14.92 & 44.31 & 22.13 & - & - & 20.73 & - & - & 2.94 & 14.59 \\
    & & & SEA & 25.32 & 19.94 & - & 11.47 & 43.27 & 10.74 & - & - & 6.88 &- & - & 3.73 & 15.04 \\
    &  & 180 & ENG & 31.04 & - & - & - & 61.92 & 33.00 & - & - & - & - & - & 3.51 & 13.76 \\
    & & & SEA & 28.22 & - & - & - & 73.53 & 14.62 & - & - & - & - & - & 4.49 & 17.39 \\
    &  & 180 & ENG & 17.67 & - & - & - & 42.78 & 22.13 & - & - & - & - & - & 4.09 & 8.59 \\
    & & & SEA & 17.14 & - & - & - & 38.73 & 13.60 & - & - & - & - & - & 3.99 & 9.25 \\
    &  & 120 & ENG & 14.72 & 11.27 & - & 16.25 & 24.04 & 12.86 & - & - & 9.85 & - & - & 9.44 & 7.38 \\
    & & & SEA & 13.97 & 10.06 & - & 14.99 & 22.51 & 9.20 & - & - & 7.34 & - & - & 3.01 & 7.83 \\
    &  & 30 & ENG & 5.12 & 4.77 & 5.48 & 20.67 & 14.00 & 4.74 & 5.56 & 8.74 & 5.49 & 6.76 & 7.32 & 2.41 & 5.06 \\
    & & & SEA & 4.79 & 4.61 & 5.28 & 12.68 & 10.88 & 4.66 & 5.39 & 5.97 & 5.99 & 6.87 & 7.70 & 2.47 & 5.43 \\
    &  & 60 & ENG & 9.48 & 8.21 & - & 14.90 & 19.07 & 7.74 & - & 9.20 & 6.58 & 7.38 & 7.49 & 2.64 & 6.82 \\
    & & & SEA & 9.08 & 7.54 & - & 11.70 & 14.72 & 6.67 & - & 7.28 & 6.71 & 7.22 & 7.42 & 2.72 & 7.69 \\
    \bottomrule
    \end{tabular}}
\end{table*}

\clearpage
\begin{table*}[!ht]
    \centering
    \caption{TLoc performance: results across datasets comparing English and SEA prompts. Evaluation is based on the F1 score metric, where higher values indicate better performance.}
    \label{tab:appendix_tloc}
    \resizebox{0.99\linewidth}{!}{%
    \renewcommand{\arraystretch}{1.5}
    \begin{tabular}{
        c@{\;}c@{\;}c@{\;}c@{\;}|
        *{12}{c@{\hspace{0.3em}}}
        c
        }
    \toprule
    \multicolumn{4}{c|}{\textbf{Model}}
    & \makecell{\textbf{MERaLiON2} \\ \textbf{10B}} 
    & \makecell{\textbf{MERaLiON2} \\ \textbf{3B}} 
    & \makecell{\textbf{SeaLLMs} \\ \textbf{Audio} \\ \textbf{7B}}
    & \makecell{\textbf{Phi-4} \\ \textbf{multi-} \\ \textbf{-modal} \\ \textbf{instruct}}
    & \makecell{\textbf{Kimi} \\ \textbf{Audio}} 
    & \makecell{\textbf{Voxtral} \\ \textbf{mini}} 
    & \makecell{\textbf{Qwen2} \\ \textbf{Audio} \\ \textbf{7B} \\ \textbf{Instruct}} 
    & \makecell{\textbf{Qwen} \\ \textbf{2.5} \\ \textbf{Omni} \\ \textbf{3B}}
    & \makecell{\textbf{Qwen} \\ \textbf{2.5} \\ \textbf{Omni} \\ \textbf{7B}}
    & \makecell{\textbf{gemma} \\ \textbf{3n} \\ \textbf{E4B-it}}
    & \makecell{\textbf{gemma} \\ \textbf{3n} \\ \textbf{E2B-it}}
    & \makecell{\textbf{Gemini} \\ \textbf{2.5} \\ \textbf{Flash}} 
    & \makecell{\textbf{GPT} \\ \textbf{4o} \\ \textbf{Audio}} \\ 
    \multicolumn{4}{c|}{Size} & 10B & 3B & 7B & 5.6B & 7B & 3B & 7B & 3B & 7B & 4B & 2B & - & - \\ 
    \midrule
    \textbf{Data} 
    & \textbf{Lang}
    & \textbf{Length (Max)}
    & \textbf{Prompt} 
    & & & & & & & & & & & & & \\ 
    \cmidrule{1-4}
    \bf ASR-SgpCCSC & zh & 30 & ENG & 11.10 & 10.85 & 8.45 & 19.90 & - & - & 19.10 & 13.10 & 31.21 & 7.86 & 6.44 & 11.89 & 22.19 \\
    & & & SEA & 11.10 & 10.85 & 8.45 & 19.90 & - & - & 19.10 & 13.10 & 31.21 & 7.86 & 6.44 & 11.89 & 22.19 \\
    &  & 60 & ENG & 6.23 & 6.42 & - & 7.73 & - & - & - & 10.78 & 12.71 & 6.64 & 2.88 & 1.26 & 10.50 \\
    & & & SEA & 6.23 & 6.42 & - & 7.73 & - & - & - & 10.78 & 12.71 & 6.64 & 2.88 & 1.26 & 10.50 \\
    &  & 120 & ENG & 4.18 & 2.59 & - & 4.66 & - & - & - & - & 7.84 & - & - & 4.55 & 6.14 \\
    & & & SEA & 4.18 & 2.59 & - & 4.66 & - & - & - & - & 7.84 & - & - & 4.55 & 6.14 \\
    \bf SG Streets & en & 30 & ENG & 40.28 & 33.56 & 22.56 & 11.96 & 17.17 & 27.70 & 54.85 & 49.61 & 44.77 & 18.04 & 19.18 & 16.62 & 40.25 \\
    & & & SEA & 40.28 & 33.56 & 22.56 & 11.96 & 17.17 & 27.70 & 54.85 & 49.61 & 44.77 & 18.04 & 19.18 & 16.62 & 40.25 \\
    &  & 60 & ENG & 23.75 & 15.16 & - & 5.30 & 11.13 & 17.90 & - & 29.78 & 38.05 & 7.36 & 12.87 & 17.90 & 37.94 \\
    & & & SEA & 23.75 & 15.16 & - & 5.30 & 11.13 & 17.90 & - & 29.78 & 38.05 & 7.36 & 12.87 & 17.90 & 37.94 \\
    \bf YODAS2 & en & 30 & ENG & 22.99 & 20.00 & 9.92 & 12.70 & 14.78 & 15.46 & 41.62 & 34.01 & 36.27 & 16.49 & 16.35 & 5.88 & 24.30 \\
    & & & SEA & 22.99 & 20.00 & 9.92 & 12.70 & 14.78 & 15.46 & 41.62 & 34.01 & 36.27 & 16.49 & 16.35 & 5.88 & 24.30 \\
    &  & 60 & ENG & 13.91 & 11.46 & - & 6.79 & 8.43 & 7.76 & - & 21.07 & 18.67 & 10.44 & 11.36 & 7.98 & 17.14 \\
    & & & SEA & 13.91 & 11.46 & - & 6.79 & 8.43 & 7.76 & - & 21.07 & 18.67 & 10.44 & 11.36 & 7.98 & 17.14 \\
    &  & 120 & ENG & 8.28 & 6.12 & - & 3.47 & 2.72 & 2.62 & - & - & 10.78 & - & - & 2.99 & 7.65 \\
    & & & SEA & 8.28 & 6.12 & - & 3.47 & 2.72 & 2.62 & - & - & 10.78 & - & - & 2.99 & 7.65 \\
    & id & 30 & ENG & 27.44 & 24.58 & 16.41 & 14.02 & 17.06 & 20.90 & 38.26 & 40.91 & 45.27 & 15.55 & 12.60 & 11.83 & 28.48 \\
    & & & SEA & 26.70 & 18.43 & 19.78 & 10.63 & 15.04 & 18.26 & 37.56 & 21.84 & 34.71 & 20.16 & 16.19 & 16.19 & 30.01 \\
    &  & 60 & ENG & 14.90 & 12.99 & - & 7.47 & 9.94 & 11.76 & - & 23.78 & 21.44 & 10.50 & 9.66 & 11.04 & 20.84 \\
    & & & SEA & 14.82 & 10.85 & - & 7.66 & 8.15 & 10.77 & - & 12.86 & 0.00 & 14.53 & 11.59 & 15.57 & 22.20 \\
    &  & 120 & ENG & 10.13 & 8.58 & - & 4.96 & 5.05 & 5.47 & - & - & 16.59 & - & - & 10.01 & 15.04 \\
    & & & SEA & 10.43 & 6.45 & - & 5.18 & 4.91 & 4.17 & - & - & 13.73 & - & - & 9.42 & 15.60 \\
    & th & 30 & ENG & 17.79 & 17.42 & 8.94 & 10.48 & 12.97 & 14.24 & 16.47 & 25.52 & 37.62 & 12.45 & 9.48 & 12.83 & 26.66 \\
    & & & SEA & 16.04 & 15.32 & 9.16 & 13.75 & 13.02 & 10.84 & 11.92 & 14.08 & 27.42 & 10.71 & 6.93 & 25.73 & 17.62 \\
    &  & 60 & ENG & 9.03 & 8.56 & - & 5.90 & 8.03 & 7.16 & - & 14.76 & 15.22 & 9.28 & 7.72 & 4.35 & 18.64 \\
    & & & SEA & 9.92 & 5.08 & - & 7.48 & 5.87 & 6.73 & - & 9.77 & 17.20 & 7.36 & 5.62 & 16.38 & 15.51 \\
    &  & 120 & ENG & 5.65 & 4.13 & - & 3.04 & 4.08 & 3.78 & - & - & 11.17 & - & - & 8.16 & 11.31 \\
    & & & SEA & 5.87 & 1.94 & - & 3.70 & 5.80 & 3.00 & - & - & 6.75 & - & - & 15.81 & 9.18 \\
    & vi & 30 & ENG & 22.40 & 14.50 & 8.22 & 10.30 & 15.84 & 18.81 & 35.08 & 33.18 & 39.17 & 12.14 & 10.35 & 10.11 & 25.67 \\
    & & & SEA & 24.76 & 13.49 & 14.04 & 14.63 & 8.16 & 15.44 & 28.60 & 31.59 & 40.73 & 12.23 & 10.82 & 8.70 & 28.80 \\
    &  & 60 & ENG & 12.18 & 11.54 & - & 4.29 & 9.76 & 8.86 & - & 17.98 & 24.73 & 9.35 & 9.14 & 6.35 & 11.98 \\
    & & & SEA & 14.77 & 7.92 & - & 7.70 & 11.35 & 7.30 & - & 17.96 & 28.44 & 10.95 & 8.88 & 7.28 & 21.40 \\
    &  & 120 & ENG & 6.34 & 7.34 & - & 2.44 & 4.84 & 3.76 & - & - & 12.73 & - & - & 2.83 & 7.40 \\
    & & & SEA & 6.88 & 6.76 & - & 3.11 & 1.20 & 4.42 & - & - & 20.39 & - & - & 3.40 & 12.68 \\
    & zh & 30 & ENG & 13.56 & 10.80 & 6.49 & 11.41 & 9.13 & 9.97 & 27.69 & 17.10 & 15.84 & 10.19 & 8.32 & 12.43 & 17.77 \\
    & & & SEA & 12.52 & 10.61 & 6.69 & 11.02 & 11.41 & 11.14 & 26.21 & 13.93 & 16.46 & 12.36 & 7.29 & 10.11 & 14.80 \\
    &  & 60 & ENG & 6.61 & 5.85 & - & 6.28 & 4.36 & 5.76 & - & 9.35 & 9.07 & 7.45 & 4.29 & 5.02 & 7.73 \\
    & & & SEA & 6.80 & 7.91 & - & 5.62 & 5.80 & 5.61 & - & 8.50 & 11.54 & 6.95 & 7.15 & 4.13 & 4.64 \\
    &  & 120 & ENG & 3.85 & 2.08 & - & 3.26 & 1.81 & 2.93 & - & - & 8.81 & - & - & 4.89 & 3.64 \\
    & & & SEA & 3.98 & 4.38 & - & 3.16 & 0.00 & 2.35 & - & - & 4.77 & - & - & 6.32 & 4.34 \\
    \bottomrule
    \end{tabular}}
\end{table*}

\label{sec:appendix}

\end{document}